\documentclass[runningheads]{llncs}
\usepackage{graphicx, subcaption}
\usepackage{comment}
\usepackage{amsmath,amssymb} 
\usepackage{color}
\usepackage[bottom]{footmisc}
\long\def\ignorethis#1{}

\usepackage{color}
\definecolor{gray}{rgb}{0.35,0.35,0.35}
\definecolor{red}{rgb}{1,0,0}
\definecolor{dark-green}{rgb}{0,0.4,0}
\definecolor{dark-pink}{rgb}{0.78,0.08,0.52}
\definecolor{blue}{rgb}{0,0,1}
\definecolor{orange}{rgb}{1,0.55,0}
\definecolor{white}{rgb}{1,1,1}
\definecolor{black}{rgb}{1,1,1}
\definecolor{dark-brown}{rgb}{0.2,0.1,0}

\ifdefined\ShowNotes

\else
  \newcommand{\colornote}[3]{}
\fi

\newbox\jsavebox

\begin{document}
\pagestyle{headings}
\mainmatter
\def\ECCVSubNumber{2866}  

\title{Deep Image Compression using Decoder Side Information} 


\titlerunning{Deep Image Compression using Decoder Side Information}
%
\author{Sharon Ayzik \and Shai Avidan}
\authorrunning{S. Ayzik and S. Avidan}
%
\institute{Dept. of Electrical Engineering\\ 
Tel Aviv University \\
\email{ayziksha@mail.tau.ac.il , avidan@eng.tau.ac.il}}
\maketitle
\begin{figure}[!b]
\begin{center}
\begin{tabular}{ c c }
\textbf{Without SI} 0.03199 bpp & \textbf{With SI (ours)} 0.03019 bpp\\
\includegraphics[width=0.5\linewidth]{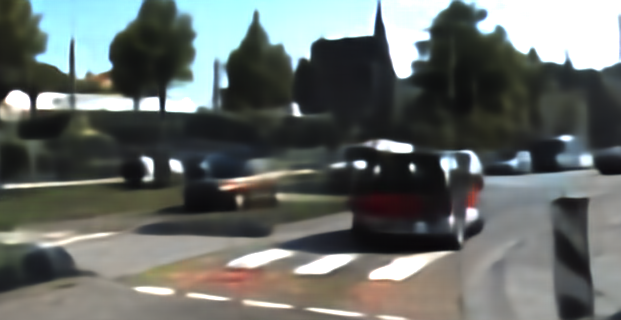} &
\includegraphics[width=0.5\linewidth]{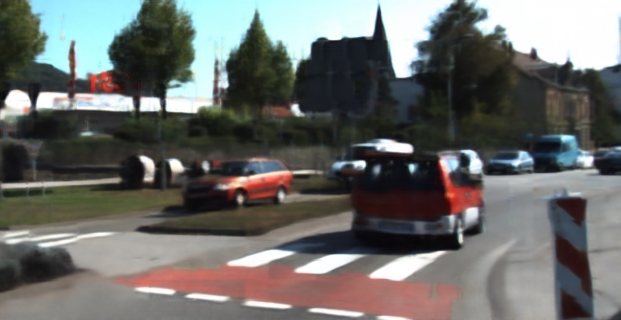} \\
\end{tabular}
\caption{Reconstruction from very low bits per pixel (bpp). Our method that use an additional Side Information (SI) image in the decoders' side can restore fine details as well as colors and textures that vanished as a result of the aggressive compression rate. Note the small red car, the crosswalk, the building to the back right side with the blue vehicle, and even the trees textures.}
\label{fig:teaser}
\end{center}
\end{figure}

\begin{abstract}

We present a Deep Image Compression neural network that relies on side information, which is only available to the decoder. We base our algorithm on the assumption that the image available to the encoder and the image available to the decoder are correlated, and we let the network learn these correlations in the training phase.

Then, at run time, the encoder side encodes the input image without knowing anything about the decoder side image and sends it to the decoder. The decoder then uses the encoded input image and the side information image to reconstruct the original image.

This problem is known as Distributed Source Coding (DSC) in Information Theory, and we discuss several use cases for this technology. We compare our algorithm to several image compression algorithms and show that adding decoder-only side information does indeed improve results. 

Our code is publicly available~\footnote{Our code is available at: \url{https://github.com/ayziksha/DSIN}}.

\end{abstract}

\section{Introduction}\label{sec:intro}

Deep Image Compression uses Deep Neural Networks (DNN) for image compression. Instead of relying on handcrafted representations to capture natural image statistics, DNN methods learn this representation directly from the data. Recent results show that indeed they perform better than traditional methods.

Ultimately, there is a limit to the compression rate of all methods, that is governed by the rate-distortion curve. This curve determines, for any given rate, what is the minimal amount of distortion that we must pay. We can break this barrier by introducing side information that can assist the network in compressing the target image even further.

Figure~\ref{fig:teaser} gives an example of results obtained by our system. The left image shows the results of a state-of-the-art deep image compression algorithm. The right image shows the results of our method that relies on side information. As can be seen, our method does a better job of restoring the details.

\begin{figure}
\begin{center}
  \includegraphics[width=0.6\linewidth]{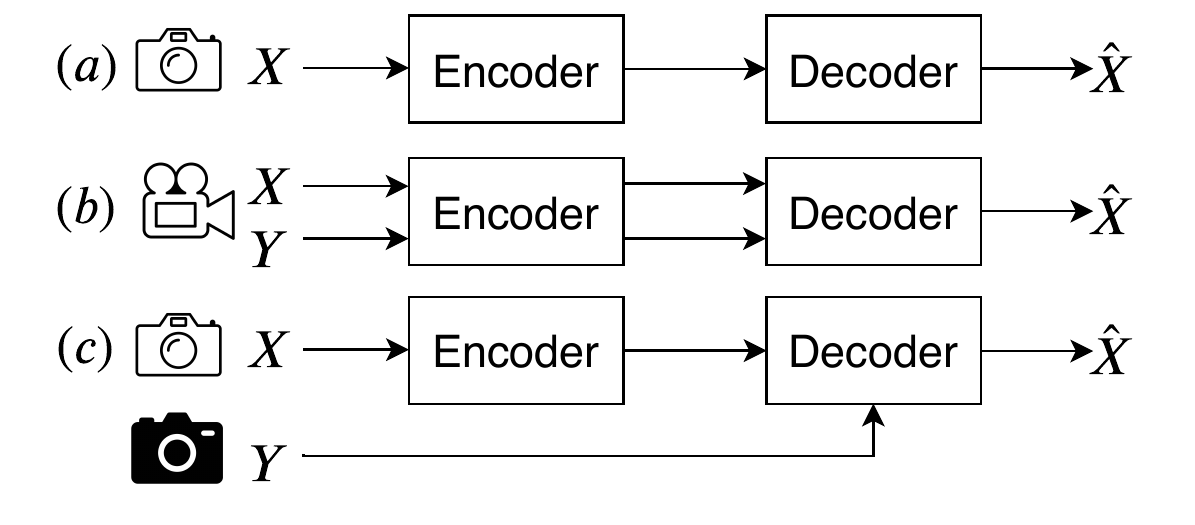}
\end{center}
  \caption{Different compression schemes. $(a)$ Single image encoding-decoding. $(b)$ Video coding: joint encoding-decoding. The successive frame $Y$ is used as side information. $(c)$ Distributed source coding -  image $X$ is encoded and then decoded using correlated side information image $Y$.}
\label{fig:enc_dec_options}
\end{figure}

One can catalogue image compression schemes into three classes (see Figure~\ref{fig:enc_dec_options}). The first (top row) is a standard image compression scheme. Such a network makes no use of side information, and the trade-off is governed by the rate-distortion curve of the image.

Deep Video Compression (second row in Figure~\ref{fig:enc_dec_options}) goes one step further and, in addition to natural image statistics, also relies on previous frames as side information that is available to both the encoder and the decoder. The availability of this side information improves the compression ratio of video compared to images. The limit of this scheme is bounded by the conditional probability of the current frame given previous frames. This works well when the two frames are correlated, as is often the case in video.

We consider a different scenario in which the side information is only available at the decoder side (third row of Figure~\ref{fig:enc_dec_options}). This is different from deep video compression, where side information is available both to the decoder and the encoder. It turns out that even in this case, the compression scheme can benefit from side information. That is, DSC can, in theory, achieve the same compression ratios as deep video compression, even though the side information is {\em not} available to the encoder. But when does this scenario occur in practice?

It turns out that this DSC scenario occurs quite frequently, and here are a couple of examples. Consider the case of a camera array. For simplicity, we focus on a stereo camera, which is the simplest of camera arrays. The left and right cameras of the stereo pair are each equipped with a micro-controller that captures the image from the camera, compresses it, and sends it to the host computer. Since both cameras capture the same scene at the same time, their content is highly correlated with each other. But since the left and right cameras do not communicate, they only communicate with the host computer and can not use the fact that they capture highly correlated images to improve the compression ratio. This puts a heavy burden on the host computer, which must capture two images in the case of stereo camera and many more in the case of a camera array.

Now suppose that the left camera transmitted its image to the host computer and the right camera as well. Then the right camera can encode its image conditioned on the left image and transmit fewer bits to the host computer. This reduces the burden on the host computer at the cost of sending the left image to the right camera. Distributed Source Coding theory tells us that we do not have to transmit the image from the left camera to the right camera at all, and still achieve the same compression ratio. When considering a camera array with multiple cameras, the savings can be substantial.

Camera arrays are assumed to be calibrated and synchronized, but we can take a much more general approach. For example, a group of people taking pictures of some event is a common occurrence nowadays. We can treat that as a distributed, uncalibrated, and unsynchronized camera array. Instead of each person uploading his images to the cloud, we can pick, at random, a reference person to upload his images to the cloud and let the rest of the people upload their images conditioned on the reference images.

Taking this idea one step further, we envision a scenario in which before uploading an image to the cloud, we will first transmit the camera's position and orientation (information that is already collected by smartphones). As a result, the cloud will be able to select existing images that are only stored in the cloud to use as side information. 

Our approach is using recent advances in deep image compression, where we add side information to the decoder side. During training, we provide the network with pairs of real-world, correlated images. The network learns to compress the input image, and then add the side information image to help restore the original image. At inference time, the encoder is used for compressing the image before transmitting it. The rest of the network, which lies at the receiver side, is used by the decoder to decode the original image, using the compressed image and the side information image. To the best of our knowledge, this is the first time Deep Learning is used for DSC in the context of image compression.

We evaluate our system on two versions of the KITTI dataset that are designed to simulate some of the scenarios described earlier. In the first, we use the KITTI Stereo dataset to simulate the scenario of a camera array (in this case, a stereo camera). In the second case, we use pairs of images from the KITTI Stereo dataset that are taken several frames apart. This case is designed to simulate the scenario where an image is uploaded to the cloud, and some other image, from the same location, is used as side information.

Our experiments show that using the side information can help reduce the communication bandwidth by anywhere between $10\%$ and $50\%$, depending on the distortion level and the correlation between the side information image and the image to be compressed.

\section{Related work}

\paragraph{\textbf{Deep compression:}}
Using DNN in many applications has gained much popularity in recent years, the same goes for the task of image compression. Common usage of DNN for the task of compression are RNNs \cite{Toderici_RNN_2016,Toderici_RNN_2017} and auto-encoders \cite{DSC4sensNetwork,balle_end_to_end_opt_img_compression,ETH}. The networks are usually designed in an end-to-end manner, aiming to minimize the final loss on the decompressed image. 

Toderici \etal \cite{Toderici_RNN_2016,Toderici_RNN_2017} used progressive image compression techniques and tested various types of recurrent neural networks to create a hybrid network that extracts a binary representation code using an entropy coder. Ball\'{e} \etal \cite{balle_end_to_end_opt_img_compression} used quantization rather than binarization. Theis \etal \cite{Lucas_Theis} use a simple approximation to replace the rounding-based quantization, in addition to bounding the discrete entropy loss. And Mentzer \etal \cite{ETH} use an auto-encoder and a context model that learns to asses the distribution of the bitstream in addition to an importance map to improve performance.

Recent work by Agustsson \etal \cite{ETH_GAN} suggests using GAN based architecture to break the rate-distortion bounds. They encode the image with fewer bits than what is dictated by the rate-distortion curve. Then they use a GAN, on the decoder side, to synthesize a similar image that is visually pleasant.

Building on the success of Deep image compression schemes, we witnessed the emergence of Deep video compression schemes. Early work replaced various steps in the video compression scheme with a DNN counterpart. For example, Lu {\em et al.} \cite{LuOXZGS18} use a deep network to remove compression artifacts in the post-processing step. Tsai {\em et al.} \cite{Tsai0S0K18} use an auto-encoder to compress the residuals of an H.264 encoder. Wu {\em et al.} \cite{WuSK18} treat video compression as a repeated image interpolation and build a full network for that. Recently, Lu {\em et al.} \cite{LuOXZGS18} proposed a network that replaces all the components of a video encoder with a single end-to-end architecture.

\paragraph{\textbf{Distributed Source Coding:}}
Distributed Source Coding (DSC) started with the groundbreaking result of Slepian-Wolf \cite{slepian_wolf,slepian_wolf_proof} who proved that it is possible to encode a source $X$ given a correlated source $Y$ even if $Y$ is only available to the decoder side. This result applied to the lossless case and was later extended by Wyner-Ziv \cite{wyner_ziv} to the lossy case by first quantizing the continuous signal and then applying the Slepian-Wolf theorem. 

Although the theory of DSC dates back to the '70s, it was only 30 years later that its first practical implementation was presented. One of the most important works was done by Pradhan and Ramchandran - Distributed Source Coding Using Syndromes (DISCUS) \cite{DISCUS}. They presented a practical framework for the asymmetric case of source coding with side information at the decoder,  based on sending the syndrome of the code-word coset for statistically dependent binary and Gaussian sources.

Much of the work on DSC was in the context of light-weight video compression. That is, instead of running a standard video compression scheme (i.e., MPEG) that requires motion estimation on the encoder side \cite{mpeg}, DSC offers the possibility of shifting the computational load from the encoder to the decoder. This scheme is useful, for example, in the case of a smartphone that needs to send a video to the cloud. For example, Girod \etal \cite{Girod_wz_video_coding,Girod_wz_video_coding_and_error_resilience,Girod_DVC} focused on Distributed Video Coding (DVC). The video sequence was split to odd and even frames, the odd frames were used as side information at the decoder while Wyner-Ziv coding was applied to the even frames.

In \cite{wz_coding_for_stereo_with_disparity,wz_multiple_imgs_unsupervised_learning_and_gray_code} the authors apply DSC to stereo images, in which one encoded image is decoded with reference to side information derived from disparity-compensated versions of the other image with the additional use of gray code.

\section{Deep Distributed Source Coding For Images}

\newcommand{\argmax}{\mathop{\mathrm{argmax}}} 
\newcommand{\argmin}{\mathop{\mathrm{argmin}}} 

\paragraph{\textbf{Toy Example:}} To gain some intuition into the DSC problem, consider the following toy example. Suppose $X$ and $Y$ are two 8-bit gray-scale images that are known to be aligned such that pixel $X(i)$ corresponds to pixel $Y(i)$. Assume image $X$ is available to the encoder on the smartphone, and image $Y$ is available to the decoder in the cloud. Transmitting $X$ to the cloud requires 8-bits per pixel. But what if the corresponding pixels, $X(i), Y(i)$ are correlated? For example, they satisfy the following correlation: $|X(i)-Y(i)| \le 3$. How can we take advantage of this correlation? A moment of thought shows that given $Y(i)$, $X(i)$ can only take seven different values, so we should hope to encode $X(i)$ using only 3 bits and not 8.

How can we do this in practice? Here is a numerical example. Let $X(i) = 110$ and $Y(i) = 113$. Consider the following DSC scheme: the encoder computes $6=\mbox{mod}(X(i), 8)$ and sends the number $6$ to the cloud using only 3 bits.  The modulo operation created a {\em coset} $\{6, 14, 22, ...,102, 110, 118, ...,254 \}$ of pixel values. Every element $x$ of this coset satisfy the constraint that $\mbox{mod}(x,8)=6$. And, by construction, the minimal distance between any pair of elements in the coset is at least $8$. Given these facts, the decoder knows that the unknown $X(i)$ must be one of the elements in the coset. It also knows that $|X(i)-Y(i)| \le 3$. Given that $Y(i)=113$, the decoder can deduce that $X(i)$ must be $110$. We have encoded $X$ using only $3$ bits per pixel, instead of $8$. Observe that the encoder did not know the value of pixel $Y(i)$ that is only available to the decoder. The prior information on the correlation between the two images is sufficient.

\paragraph{\textbf{DSC for images:}} DSC was applied to video compression, where successive frames are almost aligned. This near alignment was enough to assume that patches in successive frames that are in the same location in the image plane are correlated. Applying DSC to video compression did not get traction because the side information (i.e., previous frame in the case of video) is known to the encoder as well as the decoder. 

Here, we consider the case where the two images are taken by two {\em different} cameras at slightly different time steps. We assume that one camera uploads its image to the cloud and then let the other camera upload its image conditioned on the other image {\em without} ever having access to that image. The shift in space and time is enough to render the alignment assumption useless. We can no longer assume that the two images are aligned, nor that the layout of the images is similar. For example, we would like two images containing a house next to a tree to be correlated even if the tree is to the right of the house in one image, and is to the left in the other.

One way to address this challenge is to break the two images into patches and use patches to measure the correlation between the images. But this raises a new problem- instead of having one (image) $X$ and one (image) $Y$, we now have multiple (patches) $X$ and multiple (patches) $Y$. Now, if we use the coset trick, then we don't know which patch in $Y$ to use since the images are not aligned.

We have conflicting demands. On the one hand, we need to transmit sufficient information about a patch in $X$, to allow the decoder to pick the correlated patch in $Y$. On the other hand, we only wish to transmit a code for the coset and let the decoder use that, together with the corresponding patch in $Y$, to recover the correct patch in $X$. We solve the DSC problem for images using DNN.

\subsection{Architecture}\label{sec:Method_Arch} 
\begin{figure}[htb!]
\begin{center}
  \includegraphics[width=0.9\linewidth]{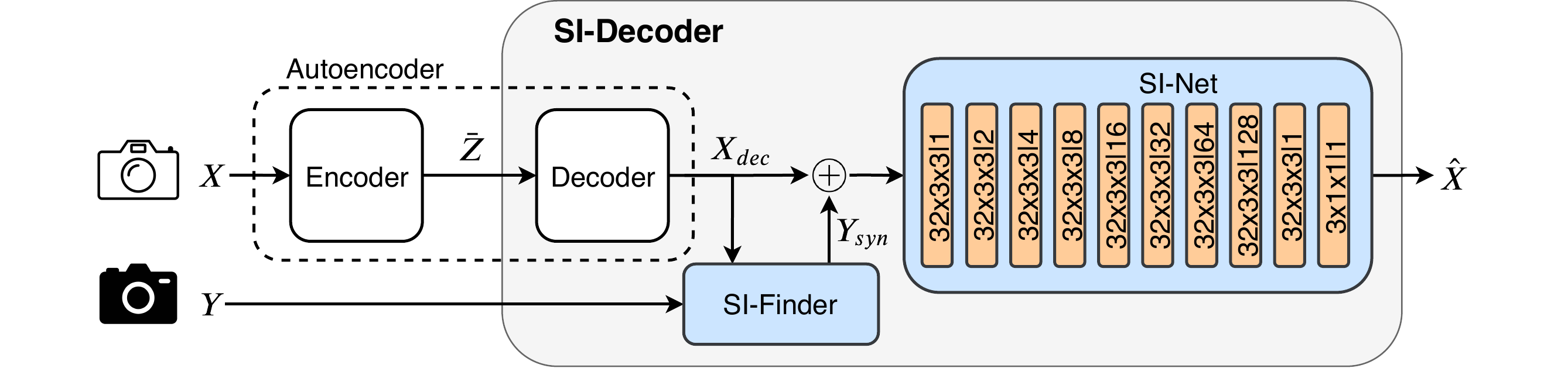}
\end{center}
  \caption{Our network's architecture. The image $X$ is encoded to ${\bar Z}$ and decoded to the image $X_{dec}$ using the auto-encoder model based on \cite{ETH}. $X_{dec}$ is used to create $Y_{syn}$ using the SI-Finder block that finds for each patch in $X_{dec}$, the closest patch in $Y$. $X_{dec}$ and $Y_{syn}$ are concatenated (marked as $\oplus$) and forwarded to the SI-Net block that outputs the final reconstruction - ${\hat X}$. The SI-Net block is based on \cite{siNet} and uses convolution layers with increasing dilation rates that approximate enlarged convolutions receptive field. $C \times K \times K$ notation in the SI-Net block refers to $K\times K$ convolutions with $C$ filters. The number following the pipe indicates the rate of kernel dilation.}
\label{fig:DDSC_arch}
\end{figure}

The overall architecture of the network is given in Figure~\ref{fig:DDSC_arch}. The encoder has access to the input image $X$, and the decoder has access to a correlated image $Y$. Our architecture consists of two sub-networks, the first is an auto-encoder designed for image compression and based on the model of Mentzer \etal \cite{ETH}. It takes the input image $X$ and produces the decoded image $X_{dec}$. The second network takes the decoded image $X_{dec}$ along with image $Y$ and uses it to construct a synthetic side information image $Y_{syn}$. The decoded image $X_{dec}$ and synthetic side information $Y_{syn}$ are then concatenated and used to produce the final output image $\hat X$. The entire network, consisting of both sub-networks, is trained jointly. Then, at inference time, the encoder uses the encoder part of the auto-encoder sub-network, while the decoder uses the rest of the network.

\begin{figure}[t]
\begin{center}
   \includegraphics[width=0.6\linewidth]{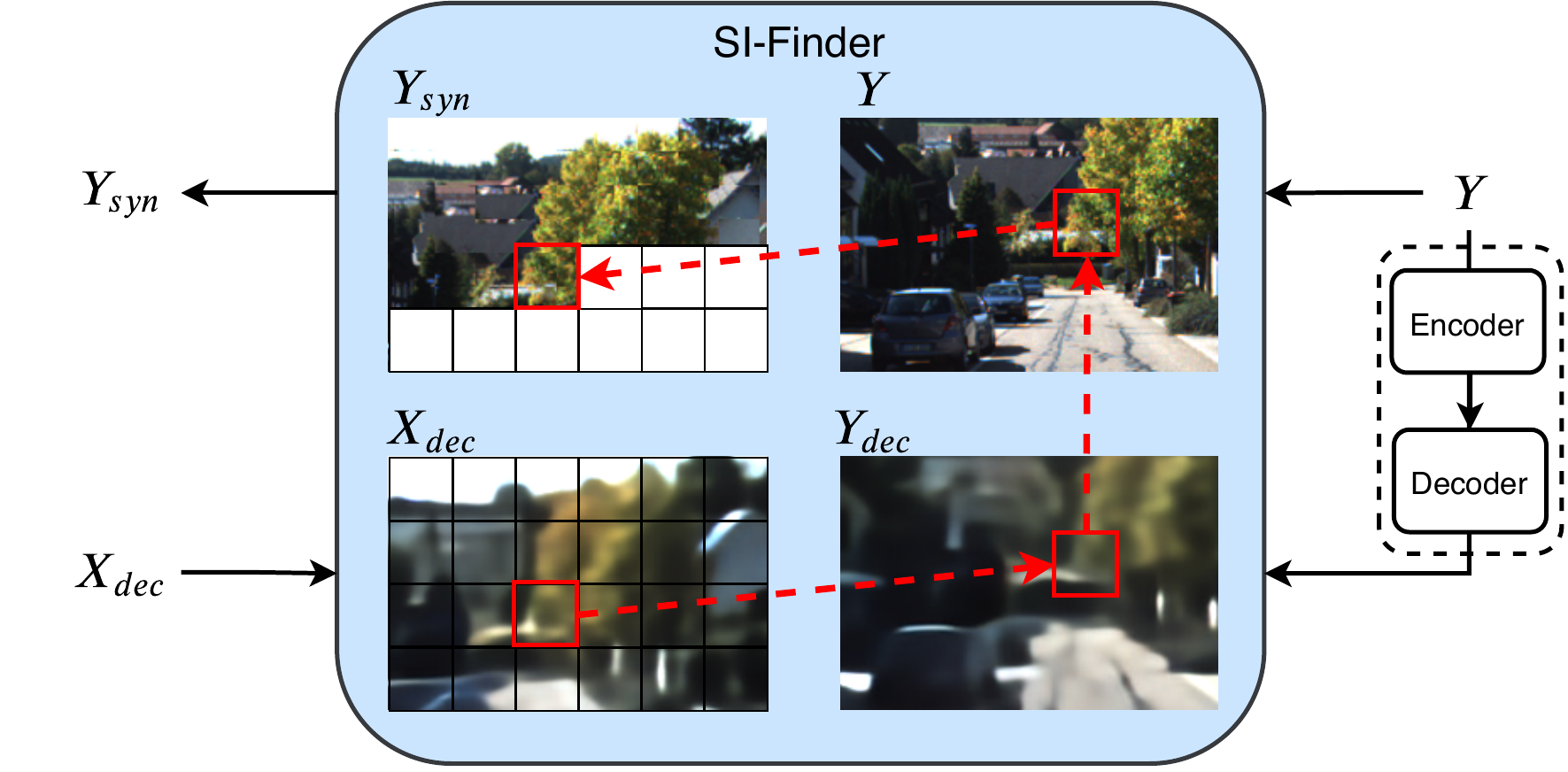}
\end{center}
  \caption{SI-Finder block illustration. This block receives $X_{dec}$ and $Y$ images, projects $Y$ to the same plane as $X_{dec}$ by passing $Y$ through the auto-encoder in inference mode to receive $Y_{dec}$. Each non-overlapping patch in image $X_{dec}$ is compared to all possible patches in $Y_{dec}$. The location of the maximum correlation patch in $Y_{dec}$ is chosen, and the corresponding patch is taken from $Y$ image. Finally, the patch is placed in $Y_{syn}$ in the corresponding $X_{dec}$ patch location.}
\label{fig:siFinder_arch}
\end{figure}

It should be noted that the quantized latent vector ${\bar Z}$ of our auto-encoder network is not designed to reconstruct the original image $X$, nor is it designed to create a coset from which the decoder can recover the correct $X$. Its goal is to provide sufficient information to construct a good synthetic image $Y$ that, together with the decoded image $X_{dec}$, can be used to recover the final result $\hat X$. This means it should reconstruct an image $X_{dec}$ that has sufficient details to search for good patches in $Y$ that are as correlated, as much as possible, with their corresponding patches in $X$.

Formally, image compression algorithms encode an input image $X$ to some quantized latent representation ${\bar Z}$ from which they can decode a reconstructed image $X_{dec}$. The goal of the compression is to minimize a distortion function. The trade-off between compression rate and distortion is defined by:
\begin{equation} 
\label{eq:RD_tradeoff}
     d(X,\hat{X}) + \beta H(\bar{Z})
\end{equation}
where $H(\bar Z)$ is the entropy of ${\bar Z}$ (i.e., the bit cost of encoding ${\bar Z}$), $d(X, {\hat X})$ is the distortion function and $\beta$ is a scalar that sets the trade-off between the two.

\subsection{Using Side Information}\label{sec:Method_Using_SI} We wish to minimize~\eqref{eq:RD_tradeoff} given a correlated image $Y$ that is only available to the decoder. To do that, we wish to create an image $Y_{syn}$ from $Y$ that is aligned with $X$. Let $f$ encode the offset of every patch in $X_{dec}$ to its corresponding patch in $Y_{dec}$, where $Y_{dec}$ is the result of passing $Y$ through the auto-encoder:
\begin{equation}
    f(i) = \argmax_{j} \mbox{corr}(\pi(X_{dec}(i)),\pi(Y_{dec}(j)))
\end{equation}

where $\mbox{corr}(\cdot)$ is a correlation metric, $\pi(X_{dec}(i))$ is the patch around pixel $X_{dec}(i)$. Then the synthetic image $Y_{syn}$ is given by:
\begin{equation}
    Y_{syn}(i) = Y(f(i))
\end{equation}
That is, $Y_{syn}$ is a reconstruction of $X$ from $Y$. We perform this reconstruction step in the SI-Finder block, which is illustrated in Figure~\ref{fig:siFinder_arch}. It receives the images $X_{dec}$ and $Y$. We then pass $Y$ through the auto-encoder to produce $Y_{dec}$ (this is only done at inference mode, so the encoder does not learn anything about $Y$). We do this since we found that matching $Y_{dec}$ with $X_{dec}$ works better than matching $Y$ with $X_{dec}$. Then, the SI-Finder compares each non-overlapping patch in $X_{dec}$ to all possible patches in $Y_{dec}$. This creates a (sparse) function $f$ that is used to create $Y_{syn}$ from $Y$. It should be noted that the SI-Finder is implemented as part of the network graph using CNN layers but is non-trainable since the CNN kernels are the image $X_{dec}$.

Eventually we feed $X_{dec}$ and $Y_{syn}$ to the SI-Net block and let it try to reconstruct $X$. Since we use concatenation of $X_{dec}$ to the side information image $Y_{syn}$ during training, we must maintain a reconstruction loss over $X_{dec}$. Therefore, the total rate-distortion trade-off from \eqref{eq:RD_tradeoff} is set to be:
\begin{equation}
\label{eq:RD_tradeoff_with_SI}
    (1 - \alpha)\cdot d(X,X_{dec}) + \alpha d(X,{\hat X})+ \beta H({\bar Z})
\end{equation}
where $\alpha$ denotes the weight for the final system's output $\hat{X}$, and the total distortion weight sums to 1 in order to maintain the balance between the distortion and the rate. 

\section{Experiments}

In the following section, we discuss the datasets we use and the training procedure in \ref{sec:impementation_details}, and then present the results of our experiments in \ref{sec:results}.
A detailed example regarding our chosen prior, images from our constructed dataset, and additional visual results appear in the supplementary material.

\subsection{Implementation details}\label{sec:impementation_details}
\paragraph{\textbf{Datasets:}}

We constructed our datasets from the KITTI 2012 \cite{KITTI2012} and KITTI 2015 \cite{dataset_KITTI2015_1,dataset_KITTI2015_2} datasets to approximate the two settings discussed in section~\ref{sec:intro}.

The first termed {\em KITTI Stereo}, consists of 1578 stereo pairs taken from the calibrated stereo cameras in the KITTI stereo datasets (\ie a pair of two images each taken at the same time from a different camera). It is designed to illustrate the calibrated and synchronized camera array use case.

The second termed {\em KITTI General}, consists of 789 scenes with 21 stereo pairs per scene taken sequentially. We constructed the dataset from pairs of images where one image is taken from the left camera and the second image from the right camera, but now, the images are taken from different time steps, in our case, 1 to 3 time steps apart. In this dataset, the images are taken up to $\sim 9$ meters apart. As a result, objects between the two images can change scale, position, or even not appear at all. This dataset is designed to simulate a much more general case where images are only loosely co-located in space or time.

\paragraph{\textbf{Evaluation criteria:}}
Following \cite{Johnston_2018_CVPR,ETH,rippel,Toderici_RNN_2016} we evaluated our results by an averaged rate-distortion curve using MS-SSIM \cite{ms-ssim}, which is reported to correlate better with human perception of distortion than mean squared error (MSE) and variants such as PSNR especially in cases where distortion is large \cite{ssim1,ssim2}.

\paragraph{\textbf{Training:}}
We implemented our model in {\em TensorFlow}. For each dataset and bit rate in the range of 0.02 to 0.2 bpp, we trained the baseline model (\ie, auto-encoder only) according to the training details in \cite{ETH}, using $L_1$ reconstruction loss (we found that training using the MS-SSIM loss for the low bit rates, suffers from instabilities and failed to reach the desired bit rates. In contrast, the $L_1$ loss led to shorter training time and better stability of the algorithm). Then we trained the full model with image size of  $320 \times 960$ for $300K$ iterations (which took around 24h per model on a single GPU), using the pre-trained auto-encoder weights as initialization and $L_1$ loss on both the auto-encoder and the final output - ${\hat X}$ with the trade-off weight $\alpha = 0.7$ from \eqref{eq:RD_tradeoff_with_SI}. We used Adam \cite{Adam} optimizer with an initial learning rate of $1\cdot10^{-4}$ and a batch size of 1 (i.e., each iteration included a pair of images $X, Y$). When training the full model, the original image size was used to enable the SI-Finder module full freedom of choice for locating the best patch possible for any given patch of $X_{dec}$. As for the SI-Finder block, we used a patch size of $20 \times 24$, and the similarity measure between patches was chosen to be the Pearson correlation on color transformed images according to \cite{RGB_to_H_transform}. We tested our models on 790 images for {\em KITTI Stereo} and 556 for {\em KITTI General} with size of $320 \times 1224$.

\subsection{Results}\label{sec:results}
We compared our baseline model without side information (i.e., auto-encoder only) to the model trained with side information. In addition, we compare ourselves to JPEG 2000 \cite{jpeg2000} and to BPG \cite{BPG} (HEVC based image codec that surpassed all other codecs in the past). For BPG, we used the non-default 4:4:4 chroma format following \cite{rippel}. We also compared our results to JPEG \cite{jpeg} and WebP \cite{WebP}, but they failed to reach our low bit rates and therefore are not shown here. We focus on low bit rates because they demonstrate the power of DSC to leverage the side information.
\begin{figure}[t]
\begin{center}
\begin{tabular}{ c c }
{\em KITTI Stereo}  &
{\em KITTI General} \\

\includegraphics[width=0.4\linewidth]{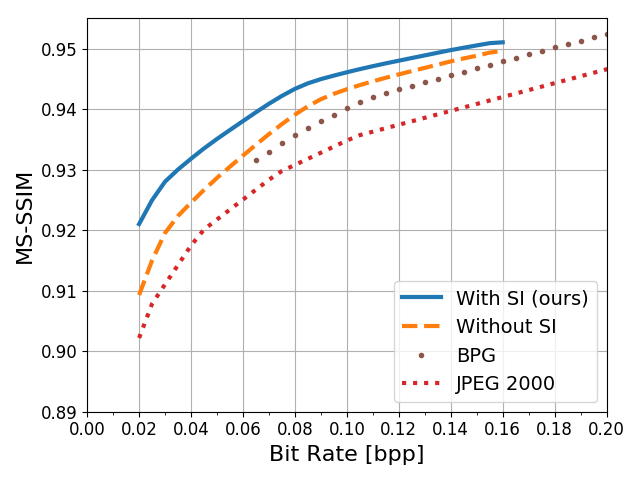} &
\includegraphics[width=0.4\linewidth]{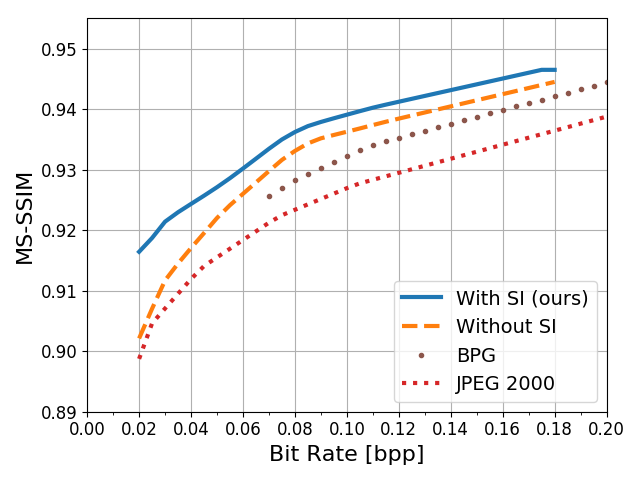}

\end{tabular}
\caption{Rate-Distortion curve - MS-SSIM as a function of bit rate on both datasets. We outperform the baseline model (without side information) as well as BPG and JPEG 2000. Note the substantial amount of bits that can be 'saved' by our method. For example, in {\em KITTI Stereo}, looking at the same value of 0.93 MS-SSIM, it can be seen that instead of sending 0.05 bpp (using the baseline model), one can send 0.03 bpp, meaning $40\%$ reduction.}
\label{fig:RD_all_no37}
\end{center}
\end{figure}
\begin{figure}[b!]
\begin{center}
\begin{tabular}{c c c c}

\textbf{JPEG 2000} & \textbf{BPG} & \textbf{Without SI} & \textbf{With SI (ours)} \\
\includegraphics[width=0.22\linewidth]{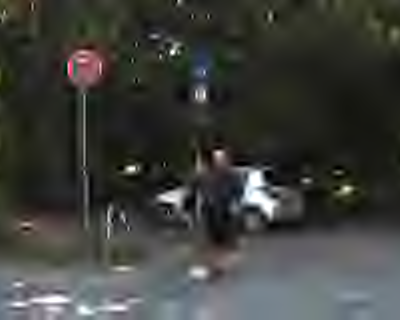} &
\includegraphics[width=0.22\linewidth]{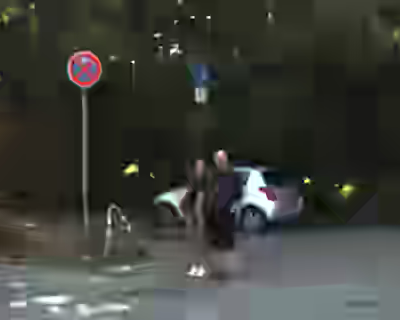} &
\includegraphics[width=0.22\linewidth]{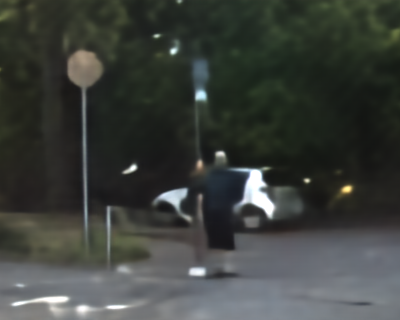} &
\includegraphics[width=0.22\linewidth]{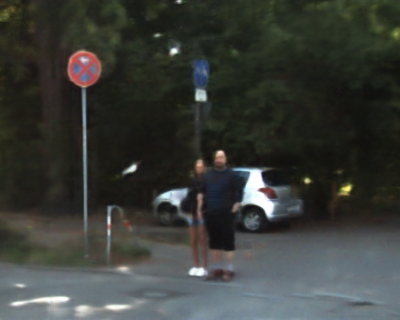} \\
0.04326 bpp & 0.04841 bpp & 0.04439 bpp & 0.04310 bpp \\

\includegraphics[width=0.22\linewidth]{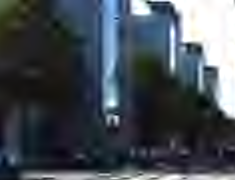} &
\includegraphics[width=0.22\linewidth]{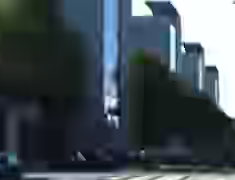} &
\includegraphics[width=0.22\linewidth]{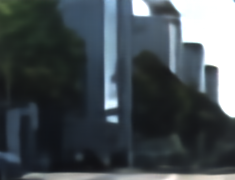} &
\includegraphics[width=0.22\linewidth]{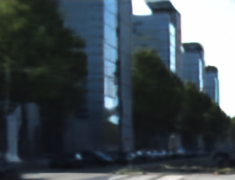} \\
0.05310 bpp & 0.05594 bpp & 0.05257 bpp & 0.05210 bpp \\

\end{tabular}
\caption{Exaamples for images compressed using the different codecs. Top row: {\em KITTI Stereo}, bottom row: {\em KITTI General}.}
\label{fig:examples_all_no37}
\end{center}
\end{figure}

Following~\cite{ETH}, and to perform a fair comparison, for each image in the test set, we extracted the sets of matching bpps and MS-SSIM measurements. Since each image has a different rate-distortion curve, we created for each image an interpolated curve and averaged the curves for all test images using a dense bpp grid. We did the same for the baseline model, BPG and JPEG 2000.

We report the results in Figure~\ref{fig:RD_all_no37}. As can be seen, our method (using side information) outperformed all compared methods. The gains are quite substantial. For example, In the {\em KITTI Stereo} dataset at MS-SSIM score of $0.94$, the bpp rate drops from about 0.08 bpp using no side information to about 0.065 bpp using side information, a drop of nearly $20\%$. The drop is even larger compared to other methods.

In addition, when comparing the two datasets, {\em KITTI Stereo} achieved greater improvement than {\em KITTI General} when using side information. This is aligned with the theory stating that the more correlated $X$ and $Y$, a more significant improvement can be achieved.

In Figure~\ref{fig:examples_all_no37}, we visually compare reconstructed images compressed using our approach to the model without side information as well as to JPEG 2000 and BPG. It can be seen that using side information improves the reconstruction - new details that were lost due to the compression are recovered as well as the color that was lost in the quantization process. 
\begin{figure}[t!]
\begin{center}
  \includegraphics[width=0.4\linewidth]{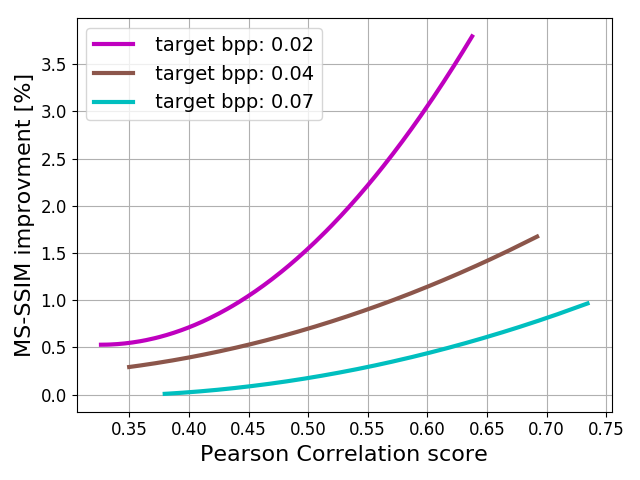}
\end{center}
  \caption{Correlation vs. Side-Information contribution on the {\em KITTI General} dataset. $x$-axis is the Pearson score between $X$ and $Y_{syn}$. $y$-axis is the improvement percentage of MS-SSIM between with and without side information models. Each curve represents a model trained to a different target bpp. As can be seen, higher correlation leads to better reconstruction results.} 
\label{fig:correlation}
\end{figure}

\paragraph{\textbf{Correlation test:}}
The DSC theorem implies that as the correlation between $X$ and $Y$ increase, so does the contribution of $Y$ in reconstructing $X$. We have seen this indirectly when analyzing the results of {\em KITTI Stereo} and {\em KITTI General}. 

To measure this connection directly, we examined the relationship between the correlation of ($X$, $Y_{syn}$) and the improvement in MS-SSIM, between the model with and model without side-information. For each test image, we calculated the average Pearson correlation between non-overlapping patches of size $(20 \times 24)$ in $X$ and their corresponding patches in $Y_{syn}$. We then computed the ratio of the MS-SSIM score of the reconstructed image $\hat X$ using the model with side-information to that without. We followed this procedure for three tested bpp rates and report the results in Figure~\ref{fig:correlation}. As expected, there is a direct link between correlation and reconstruction improvement. A higher correlation leads to better reconstruction results.
\begin{figure*}[t!]
\begin{center}
\begin{tabular}{ c c }
{\em KITTI Stereo}  &
{\em KITTI General}  \\
\includegraphics[width=0.4\linewidth]{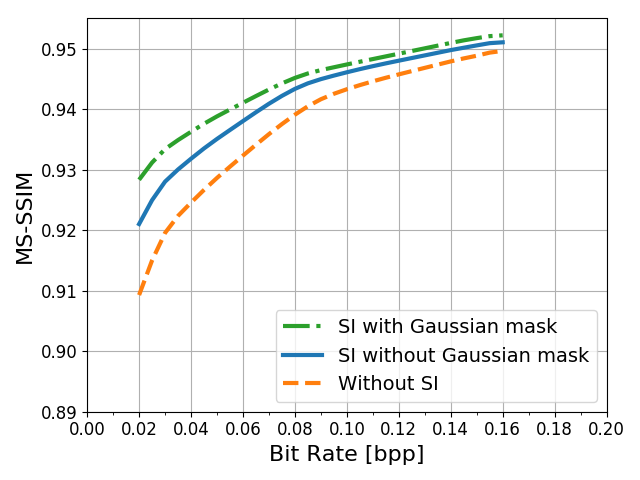} &
\includegraphics[width=0.4\linewidth]{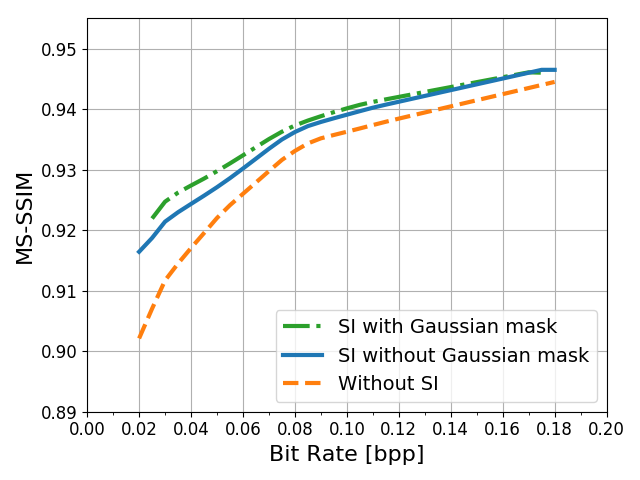}
\end{tabular}
\caption{Comparison between models trained with and without a 2D Gaussian mask. It can be seen that the results of {\em KITTI Stereo} were improved the most, as expected when using the mask.}
\label{fig:RD_no_mask_no37}
\end{center}

\end{figure*}
While all curves show positive relation, each curve (\ie, different bpp) has a different exponential growth. Therefore, providing $Y_{syn}$ with a similar Pearson score results in a more significant improvement for the lower bpp models. 
We argue that this behavior relates to the fact that Pearson score is biased towards structures,~\ie, the higher frequencies, which effected the most at low bpps. By using this metric to create $Y_{syn}$, we provide the structural information that is more beneficial in the lower bpps.

\paragraph{\textbf{Guided Search:}}
A major challenge in our technique is the matching step in which we attempt to find the correct patch in $Y$ for every patch in $X$. The results so far are based on pure visual search. However, in some cases, we might have additional information that can be used. In particular, we assume that given a patch in $X$, its corresponding patch in $Y$ should be roughly in the same location in the image plane. We enforce this assumption using a 2D Gaussian mask that helps weight patch similarity in the SI-Finder block. The mean of the mask is taken to be the position of the patch in the image plane, and the variance of the mask is roughly half image size in both axis. This encourages the SI-Finder block to pick patches in $Y$ from roughly the same location, in the image plane, as the patch in $X$.





\begin{figure}[t!]
\begin{center}
\begin{tabular}{c c c c}

Original X & Patch Match & $Y_{syn}$ without mask & $Y_{syn}$ with mask\\
\includegraphics[width=0.24\linewidth]{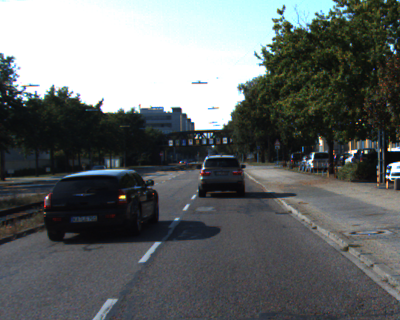} &
\includegraphics[width=0.24\linewidth]{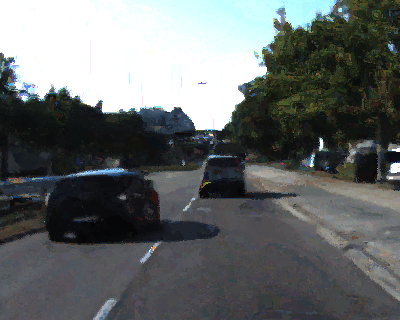} &

\includegraphics[width=0.24\linewidth]{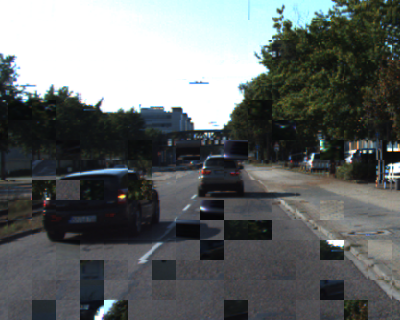} &
\includegraphics[width=0.24\linewidth]{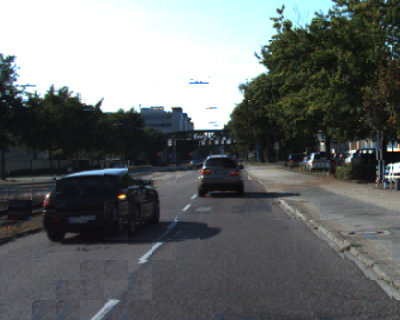} \\

\end{tabular}
\caption{Different approaches to creating $Y_{syn}$. Left to right: original $X$, PathMatch based side information, $Y_{syn}$ without and with a 2D Gaussian mask. As can be seen, using the 2D Gaussian mask as a prior improved the creation of $Y_{syn}$.}
\label{fig:no_mask}
\end{center}
\end{figure}

\begin{figure}[b!]
\begin{center}
\begin{tabular}{c c c c}

 \multicolumn{4}{c}{\textbf{Without SI}}\\
\includegraphics[height=0.18\linewidth]{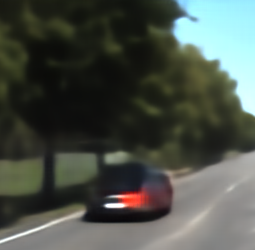} &
\includegraphics[height=0.18\linewidth]{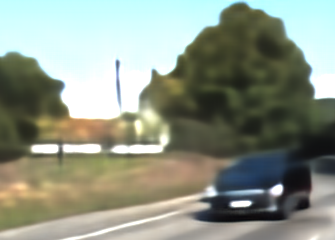} &
\includegraphics[height=0.18\linewidth]{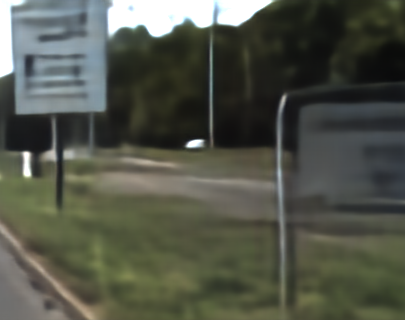} &
\includegraphics[height=0.18\linewidth]{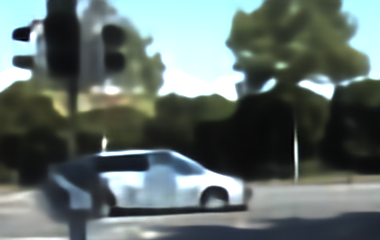} \\
0.02518 bpp & 0.03095 bpp & 0.02530 bpp & 0.03319 bpp\\
\\

 \multicolumn{4}{c}{\textbf{With SI (ours)}}\\
\includegraphics[height=0.18\linewidth]{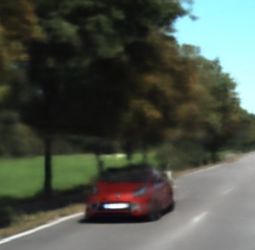} & 
\includegraphics[height=0.18\linewidth]{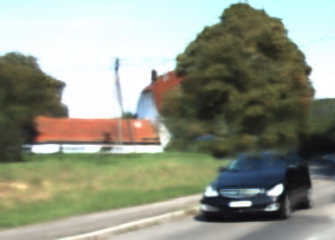} &
\includegraphics[height=0.18\linewidth]{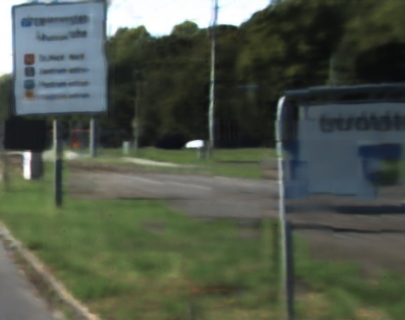} &
\includegraphics[height=0.18\linewidth]{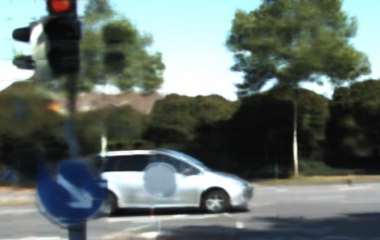} \\
0.02459 bpp & 0.02926 bpp & 0.02384 bpp & 0.03075 bpp\\
\\

\end{tabular}
\caption{Reconstruction examples from {\em KITTI Stereo} compressed using very low bpp with and without the use of side information. Complete objects, fine details and color are restored.}
\label{fig:examples_only2}
\end{center}
\end{figure}






To verify the impact of the mask on our system, we trained the full model using the SI-Finder block with and without the use of the mask. See  Figure~\ref{fig:RD_no_mask_no37}. The use of the mask improved results for most bpp (except for a single point). In Figure~\ref{fig:no_mask}, we present an example of creating $Y_{syn}$ with and without a 2D Gaussian mask. As can be seen, the mask helps the algorithm pick better patches, especially in smooth regions where the Pearson correlation score fails. For comparison, we tested PatchMatch \cite{patchmatch} to recover side information by comparing $X_{dec}$ to $Y_{dec}$ and taking the patches from $Y$. This scenario is not practical because PatchMatch does not run in the network, but it serves as a possible upper bound. As can be seen, the recovered side information using PatchMatch, in this case, looks much better. But upon close inspection, it can be seen that the high-frequency details are distorted. We tested using PatchMatch based side information and got results worst than the once reported here. Nevertheless, we leave the integration of PatchMatch into our network as a possible future research direction. In Figure~\ref{fig:examples_only2} we share additional reconstruction examples trained using the Gaussian mask and compressed to very low bit rates (that BPG failed to reach) on {\em KITTI Stereo}.

\subsection{Ablation study}\label{sec:abelation}
\begin{figure}[t!]
\begin{center}
\begin{tabular}{ c c }
{\em KITTI Stereo} & {\em KITTI General}\\
\includegraphics[width=0.4\linewidth]{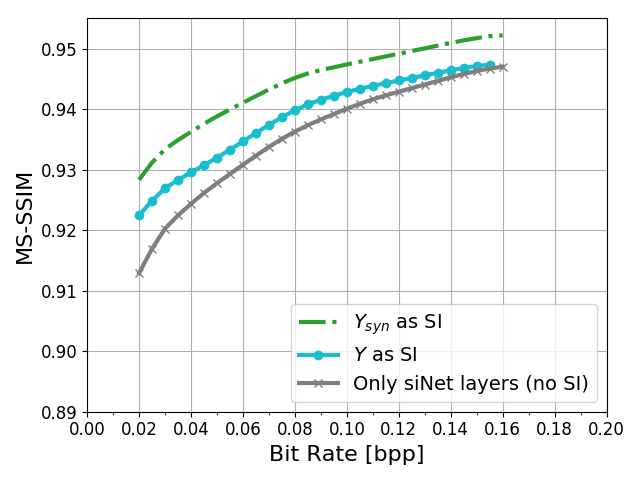} &
\includegraphics[width=0.4\linewidth]{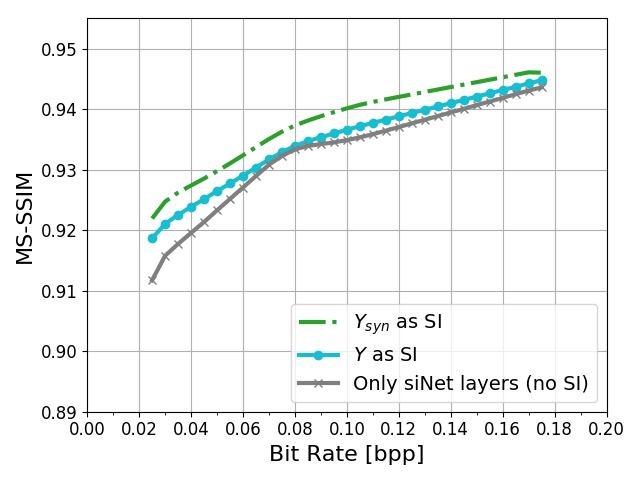}
\end{tabular}
\caption{Comparsion between models that use additional siNet layers without SI image, original $Y$ image as side information and $Y_{syn}$ (with the Gaussian mask).}
\label{fig:y_as_si_both}
\end{center}
\end{figure}

\paragraph{\textbf{Impact of SI-Net layers:}}
In order to prove that the improved reconstruction quality is a result of the side information and not an effect of the additional learnable layers, we train our network with the additional layers, \ie the SI-Net layers, but without any use of side information $Y$. As can be seen in Figure~\ref{fig:y_as_si_both}, adding layers (SI-Net block) has no effect on the reconstruction abilities of the model. Therefore, it is clear that the model gain in performance results from exploiting the additional information of the side information image.

\paragraph{\textbf{$Y$ instead of $Y_{syn}$:}}
To demonstrate the benefit of $Y_{syn}$, we trained new models (for all the bit rates) using the image $Y$ 'as is'. That is, we concatenated the image $Y$ directly with $X_{dec}$ and skipped the entire block of the SI-Finder. As can be seen in Figure~\ref{fig:y_as_si_both}, when comparing the results to the model trained with $Y_{syn}$, using $Y$ as side information yields inferior results. 


\section{Conclusions}

We proposed a novel Deep Image Compression neural network with decoder-only side information.  The proposed algorithm relies on the fact that it is possible to improve the compression of an image at the encoder, given that there is a correlated image available only at the decoder. To the best of our knowledge, we are the first to apply Deep Learning techniques to the problem of Distributed Source Coding for image compression. This scenario is quite common in practice, and we considered two such cases. The first is the case of a camera array, and the second is the case of uploading an image to the cloud, where similar images from the same location are already stored. Experiments that were designed to mimic these scenarios show that we can reduce communication bandwidth anywhere between $10\%$ to $50\%$. This demonstrates the advantages of our approach.
\newline
\newline
\textbf{Acknowledgments} This work was partly funded by ISF grant number 1549/19.

%
%

\bibliographystyle{splncs04}
\bibliography{07_references}

\clearpage

\section*{Supplementary Material}
\begin{figure}[htb!]
\begin{center}
\begin{tabular}{c}

{\em KITTI Stereo}  \\
\includegraphics[width=\linewidth]{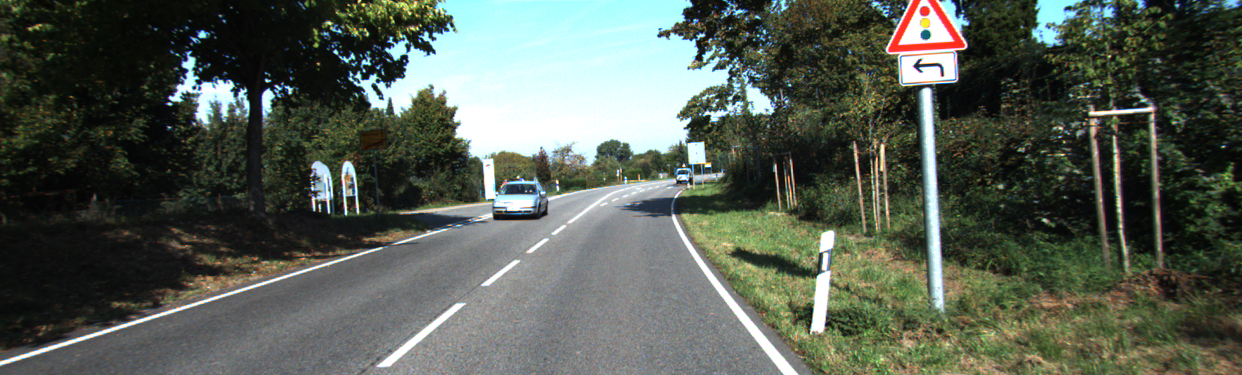} \\
\includegraphics[width=\linewidth]{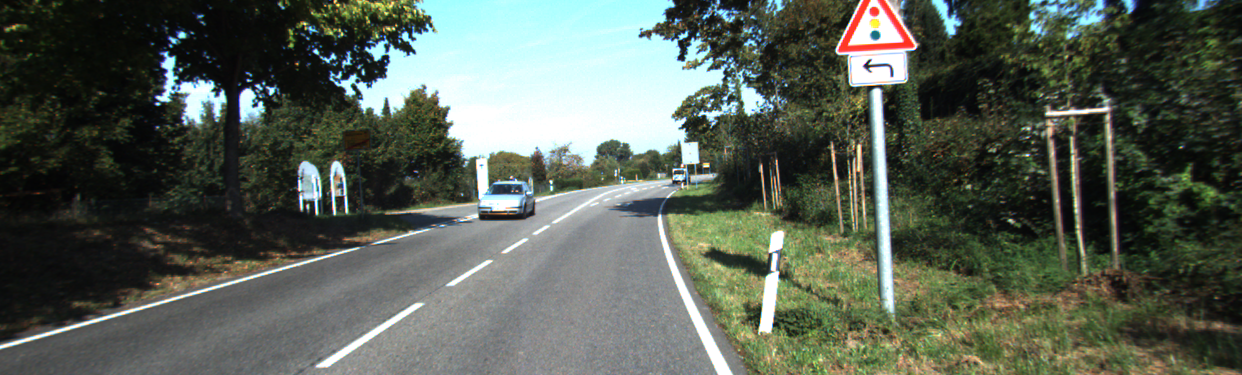} \\
\\
{\em KITTI General}  \\
\includegraphics[width=\linewidth]{Images/supplementary/dataset_example/000129_07_L.png} \\
\includegraphics[width=\linewidth]{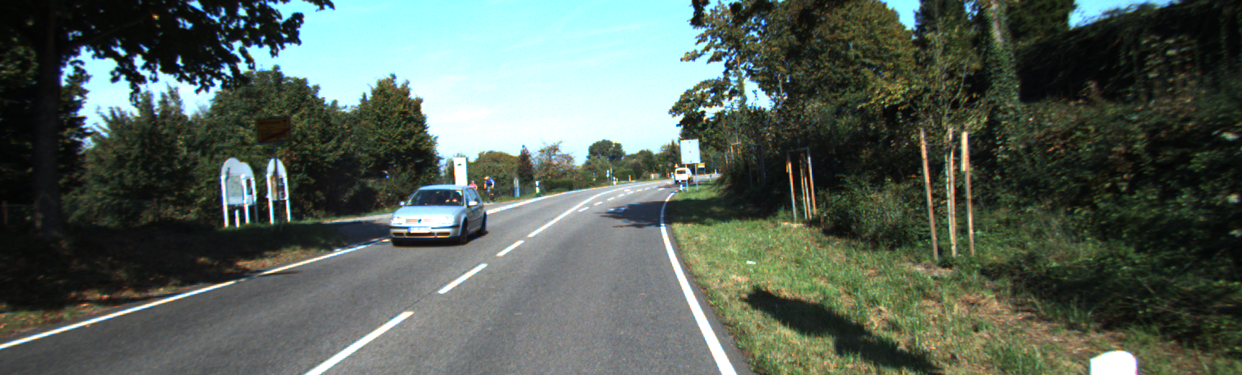} \\

\end{tabular}
\caption{Examples from both datasets. Both examples present the same $X$ (top image in each pair), while the side information $Y$ is taken according to the dataset's settings. Top pair - {\em KITTI Stereo} - the different angle between images can be seen. Bottom pair - {\em KITTI General} - in addition to the two cameras different angle, object can appear in different scale (such as the car) and some of the object are missing in the matching pair image (such as the traffic sign).}

\label{fig:supp_dataset_examples}
\end{center}
\end{figure}

\paragraph{\textbf{Dataset examples:}} Examples for both datasets can be seen in Figure~\ref{fig:supp_dataset_examples}, where the top pair is an example from {\em KITTI Stereo}, and bottom pair is an example from {\em KITTI General}. In each pair, the top image is an example for $X$ image, and below it, its matching side information image $Y$. While {\em KITTI stereo} images present the same scene from a slightly different angle, {\em KITTI General} images contain the same objects but in different scales and angles as well as objects that appear in one image but not on its matching pair.

\begin{figure*}[!b]

\begin{center}
\begin{tabular}{c c}

$X_{dec}$ & $Y_{dec}$ \\
\includegraphics[width=0.48\linewidth]{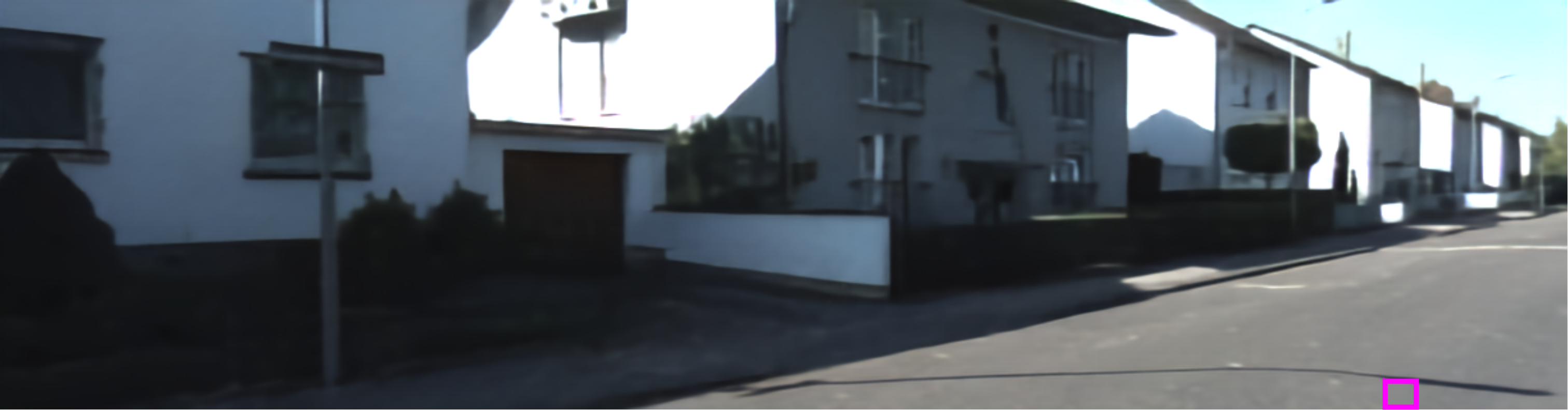} &
\includegraphics[width=0.48\linewidth]{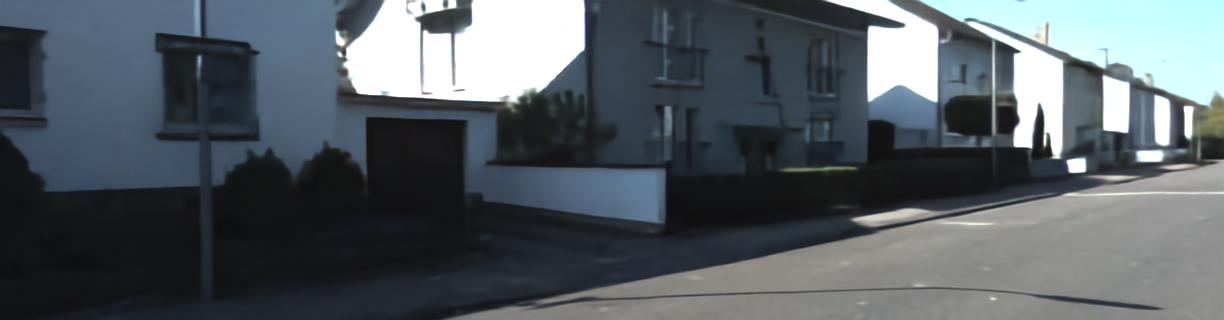} \\

Correlation map without mask & Correlation map with mask \\
\includegraphics[width=0.48\linewidth]{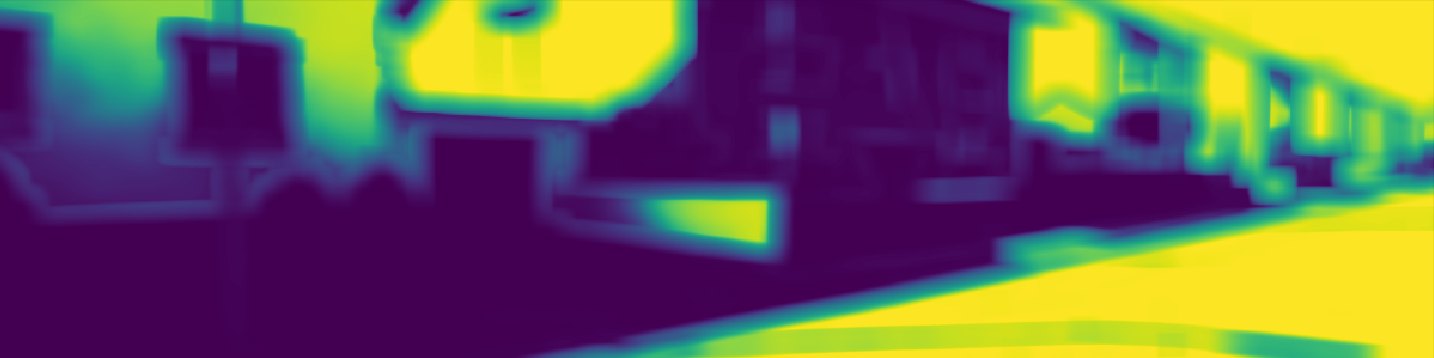} &
\includegraphics[width=0.48\linewidth]{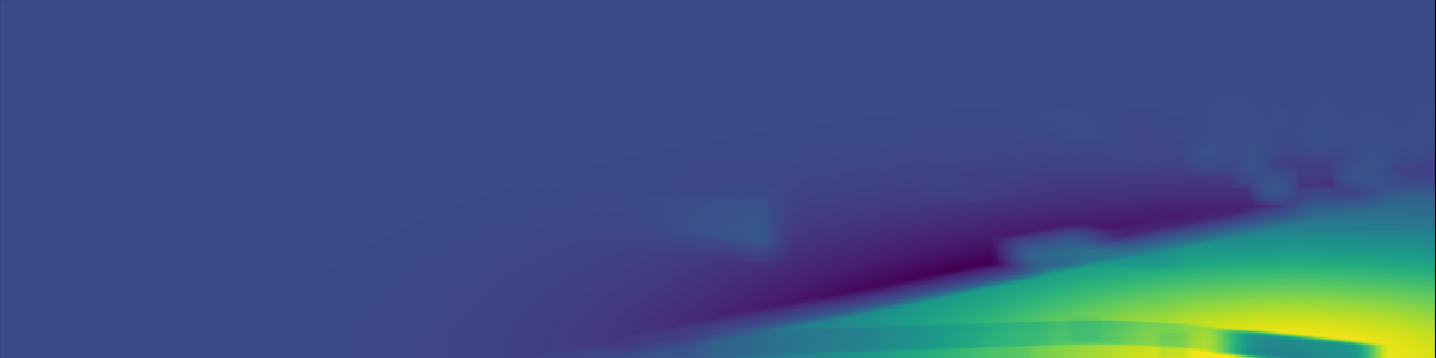} \\

\multicolumn{2}{c}{$Y$ with patch selection marked} \\
\multicolumn{2}{c}{\includegraphics[width=0.98\linewidth]{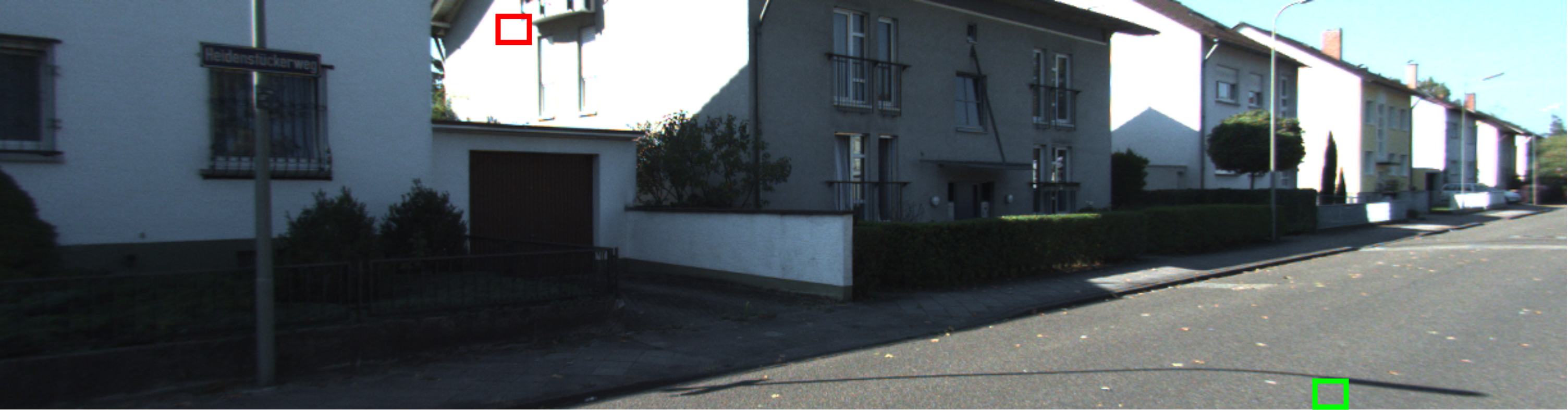}} \\

$Y_{syn}$ without mask & $Y_{syn}$ with mask \\
\includegraphics[width=0.48\linewidth]{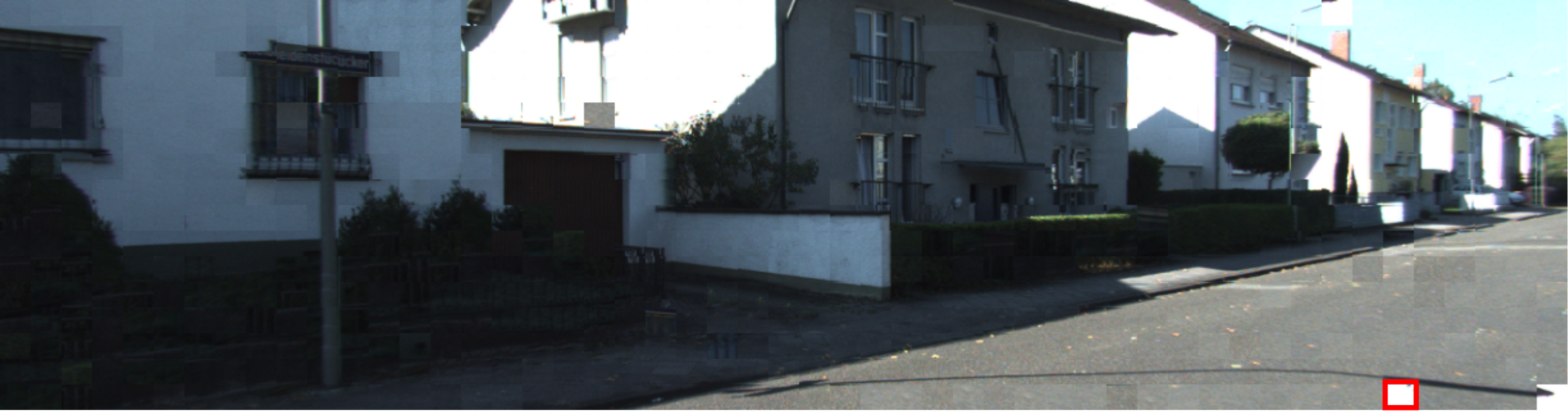} &
\includegraphics[width=0.48\linewidth]{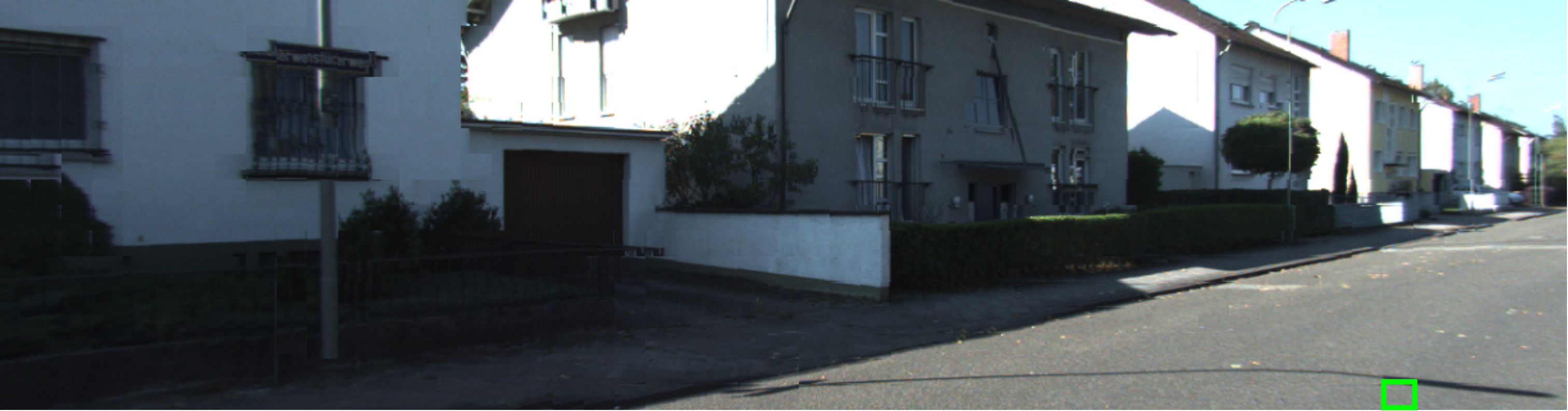} \\

\end{tabular}
\caption{Top to bottom left to right: $X_{dec}$ image with target patch marked in magenta that is compared to all possible patches in $Y_{dec}$ image (top right) and the output, is the correlation map (second row) with and without 2D Gaussian mask (yellow equals high correlation). Third row, $Y$ image with maximum score patches marked (green patch - when using the mask, red patch - without the mask). Bottom, $Y_{syn}$ image created with and without the use of the 2D Gaussian mask (the matching patches marked - red patch selected without the mask, green patch selected when using the 2D Gaussian mask).}

\label{fig:correlation_maps}
\end{center}
\end{figure*}

\paragraph{\textbf{2D Gaussian mask:}}
As mentioned in the main paper's ablation study,
we found it beneficial to add a 2D Gaussian mask as a prior in the process of creating $Y_{syn}$. In Figure~\ref{fig:correlation_maps} we present an example for a correlation map created for a certain patch of $X_{dec}$ by following our method of patch selection as mentioned in the main paper.
Furthermore, we compare correlation maps created with and without the use of a 2D Gaussian mask and show their selected matching patches (i.e., patches that yield maximum correlation) marked in $Y$ image. We can see that the 2D Gaussian mask focuses the attention on the more relevant patches.

\paragraph{\textbf{Reconstruction examples:}} 
In the next pages, we share additional visual examples for both datasets - {\em KITTI Stereo} and {\em KITTI General} compared to the baseline model, BPG, and JPEG 2000. For the other codecs, we chose the reconstruction results with the smallest bpp above ours. When very low bpp is applied, we compare ourselves only with the baseline model since BPG failed to reach these bpps. 
\newline By observing the results, we can see that JPEG 2000 yields very blurry images, while BPG restores coarse edges well but lacks in textures and fine details. Our model succeeds in restoring edges as well as fine features and textures. When comparing our model with the baseline model, our method does a better job in restoring objects, textures, and colors.  


\begin{figure}[!htb]
\begin{center}
\begin{tabular}{c }

\textbf{JPEG 2000} 0.05225 bpp \\
\includegraphics[width=\linewidth]{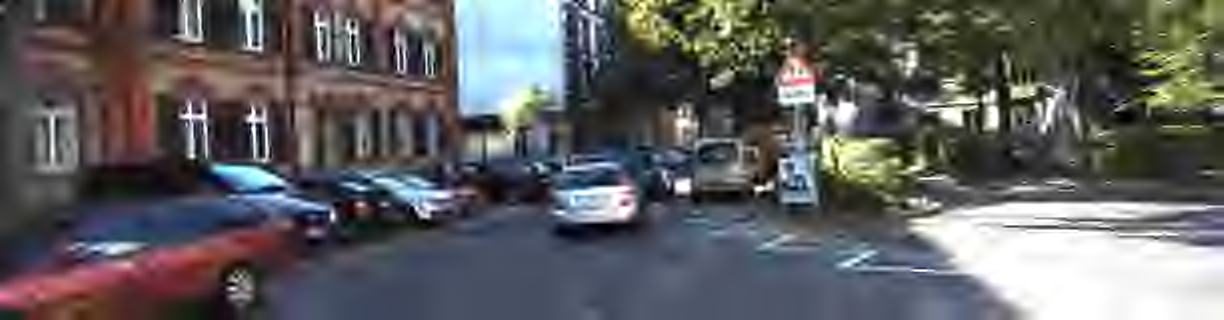} \\

\textbf{BPG} 0.05770 bpp \\
\includegraphics[width=\linewidth]{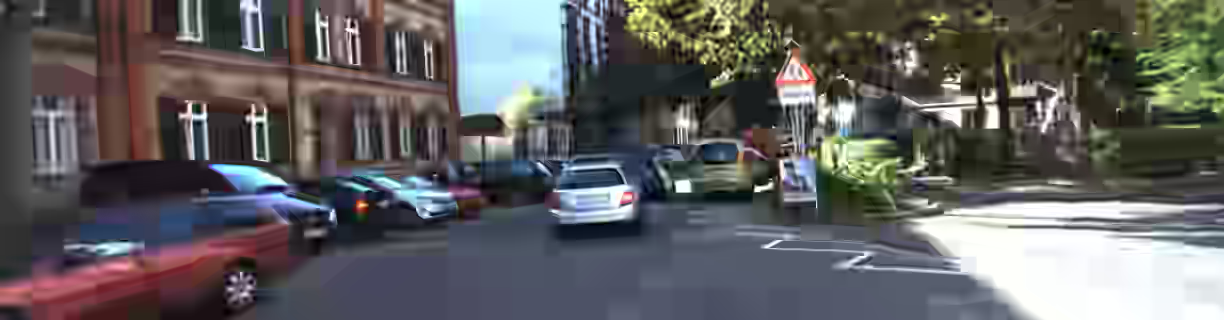} \\

\textbf{Without SI} 0.05241 bpp \\
\includegraphics[width=\linewidth]{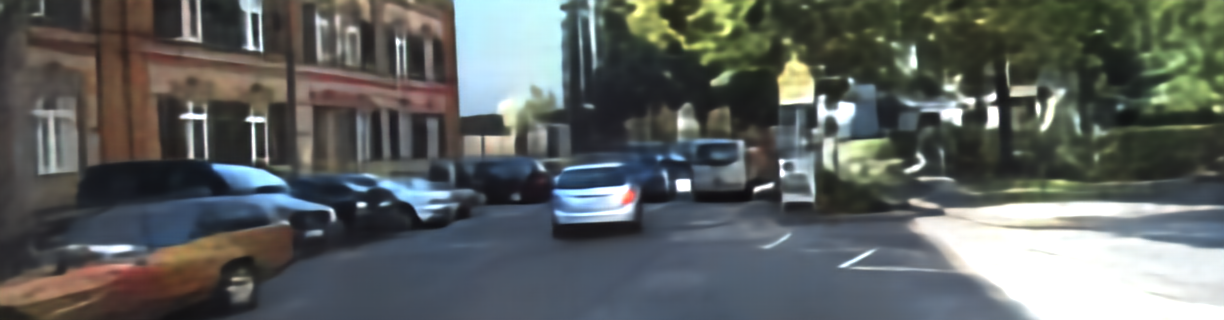} \\

\textbf{With SI} 0.05119 bpp \\
\includegraphics[width=\linewidth]{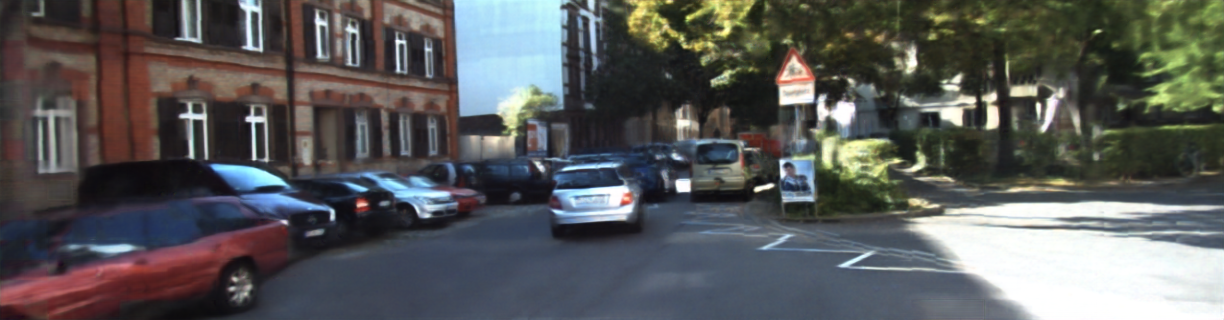} \\

\end{tabular}
\caption{Reconstruction comparison to the baseline model, JPEG 2000 and BPG over {\em KITTI Stereo}.}

\label{fig:supp_examples_stereo_fullsize_all4}
\end{center}
\end{figure}

\begin{figure*}[!htb]
\begin{center}
\begin{tabular}{c c c c}

\textbf{JPEG 2000} & \textbf{BPG} & \textbf{Without SI} & \textbf{With SI}\\

\includegraphics[width=0.2\linewidth]{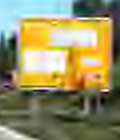} &
\includegraphics[width=0.2\linewidth]{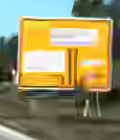} &
\includegraphics[width=0.2\linewidth]{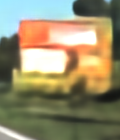} &
\includegraphics[width=0.2\linewidth]{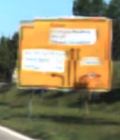} \\
0.05125 bpp & 0.04998 bpp & 0.05127 bpp & 0.04743 bpp \\


\includegraphics[width=0.2\linewidth]{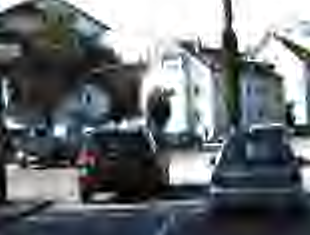} &
\includegraphics[width=0.2\linewidth]{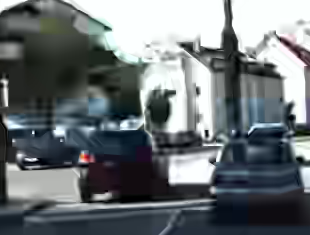} &
\includegraphics[width=0.2\linewidth]{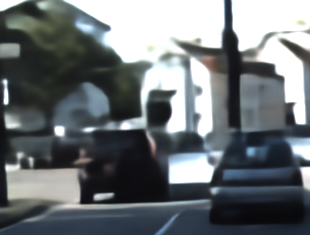} &
\includegraphics[width=0.2\linewidth]{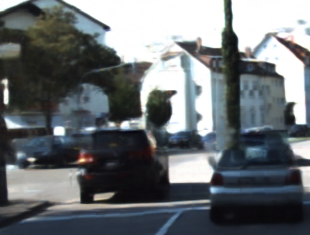} \\
0.06048 bpp & 0.05790 bpp & 0.05620 bpp & 0.05585 bpp \\


\includegraphics[width=0.2\linewidth]{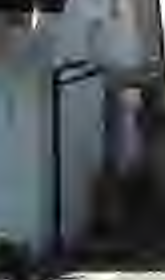} &
\includegraphics[width=0.2\linewidth]{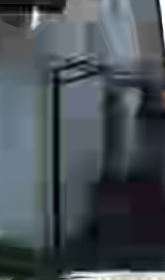} &
\includegraphics[width=0.2\linewidth]{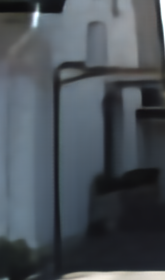} &
\includegraphics[width=0.2\linewidth]{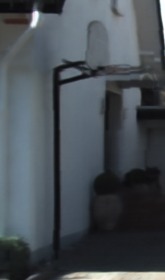} \\
0.05321 bpp & 0.05341 bpp & 0.04858 bpp & 0.04762 bpp \\


\includegraphics[width=0.2\linewidth]{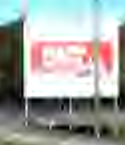} &
\includegraphics[width=0.2\linewidth]{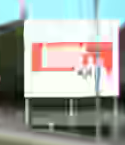} &
\includegraphics[width=0.2\linewidth]{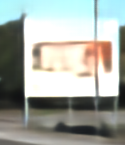} &
\includegraphics[width=0.2\linewidth]{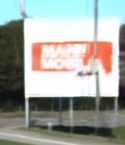} \\
0.05319 bpp & 0.05335 bpp & 0.05197 bpp & 0.04793 bpp \\


\includegraphics[width=0.2\linewidth]{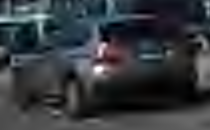} &
\includegraphics[width=0.2\linewidth]{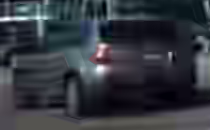} &
\includegraphics[width=0.2\linewidth]{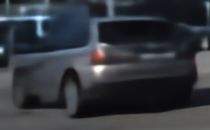} &
\includegraphics[width=0.2\linewidth]{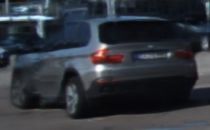} \\
0.05340 bpp & 0.05435 bpp & 0.05099 bpp & 0.04765 bpp \\

\end{tabular}
\caption{Our suggested method compared with the baseline model, JPEG 2000, and BPG over {\em KITTI Stereo}.}

\label{fig:supp_examples_stereo_all4}
\end{center}
\end{figure*}






\begin{figure*}[!htb]
\begin{center}
\begin{tabular}{c}

\textbf{JPEG 2000} 0.06319 bpp \\
\includegraphics[width=\linewidth]{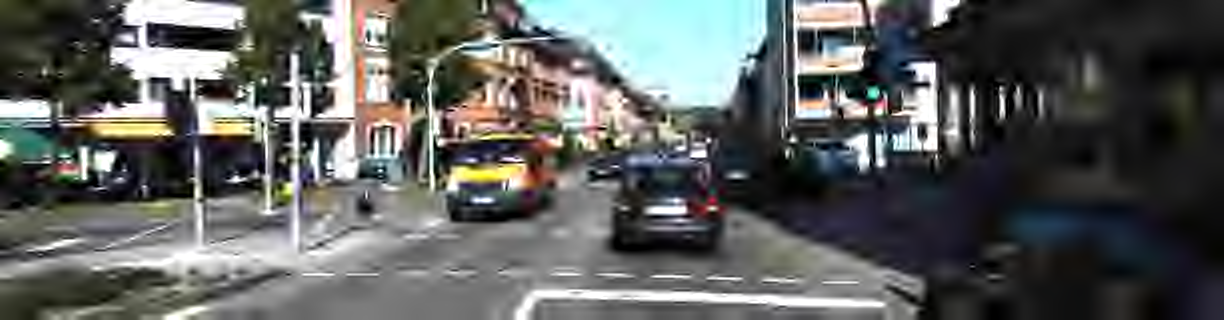} \\

\textbf{BPG} 0.06644 bpp \\
\includegraphics[width=\linewidth]{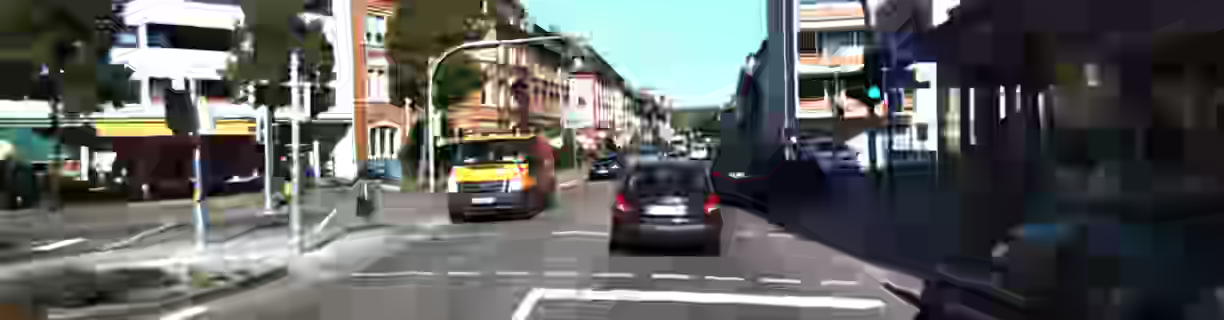} \\

\textbf{Without SI} 0.06029 bpp \\
\includegraphics[width=\linewidth]{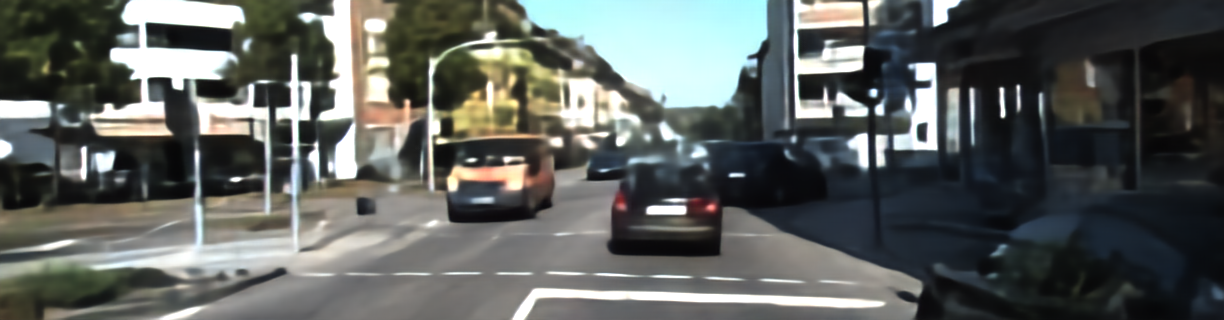} \\

\textbf{With SI} 0.05881 bpp \\
\includegraphics[width=\linewidth]{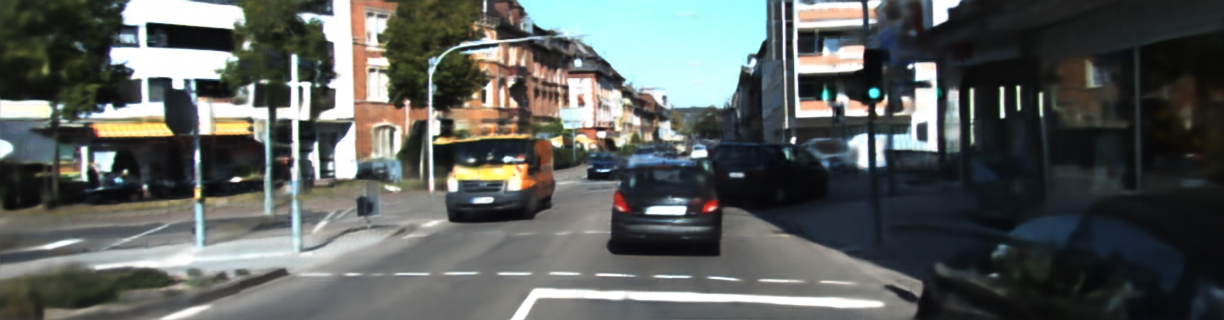} \\

\end{tabular}
\caption{Reconstruction comparison to the baseline model, JPEG 2000 and BPG over {\em KITTI General}.}

\label{fig:supp_examples_general_fullsize_all4}
\end{center}
\end{figure*}

\begin{figure*}[!htb]
\begin{center}
\begin{tabular}{c c c c}

\textbf{JPEG 2000} & \textbf{BPG} & \textbf{Without SI} & \textbf{With SI}\\
\includegraphics[width=0.2\linewidth]{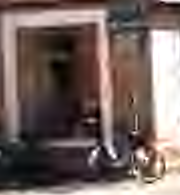} &
\includegraphics[width=0.2\linewidth]{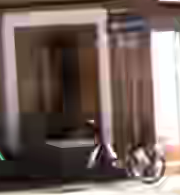} &
\includegraphics[width=0.2\linewidth]{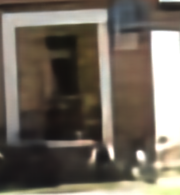} &
\includegraphics[width=0.2\linewidth]{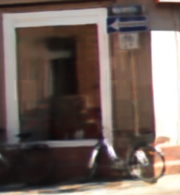} \\
0.06227 bpp & 0.06058 bpp & 0.05812 bpp & 0.05742 bpp \\


\includegraphics[width=0.2\linewidth]{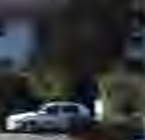} &
\includegraphics[width=0.2\linewidth]{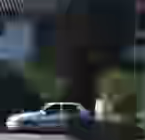} &
\includegraphics[width=0.2\linewidth]{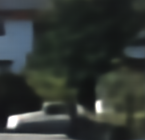} &
\includegraphics[width=0.2\linewidth]{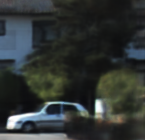} \\
0.04861 bpp & 0.04551 bpp & 0.04507 bpp & 0.04486 bpp \\


\includegraphics[width=0.2\linewidth]{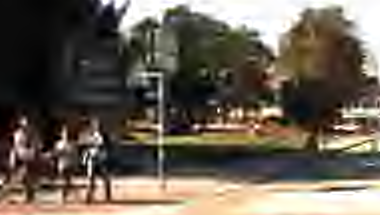} &
\includegraphics[width=0.2\linewidth]{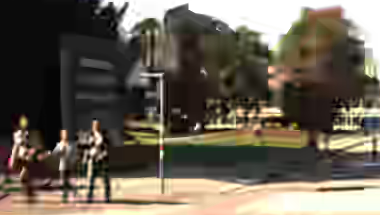} &
\includegraphics[width=0.2\linewidth]{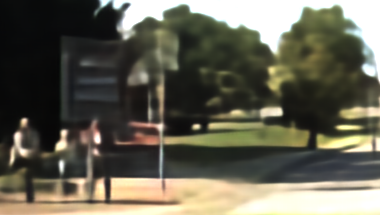} &
\includegraphics[width=0.2\linewidth]{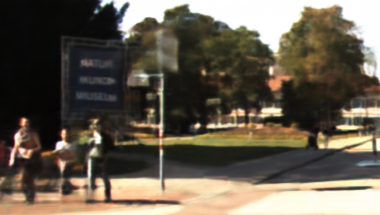} \\
0.06150 bpp & 0.06379 bpp & 0.06037 bpp & 0.05723 bpp \\


\includegraphics[width=0.2\linewidth]{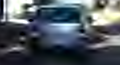} &
\includegraphics[width=0.2\linewidth]{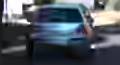} &
\includegraphics[width=0.2\linewidth]{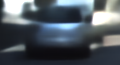} &
\includegraphics[width=0.2\linewidth]{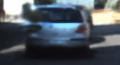} \\
0.05370 bpp & 0.05470 bpp & 0.05109 bpp & 0.04837 bpp \\


\includegraphics[width=0.2\linewidth]{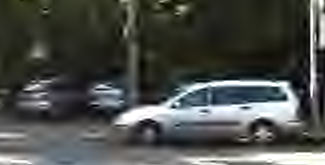} &
\includegraphics[width=0.2\linewidth]{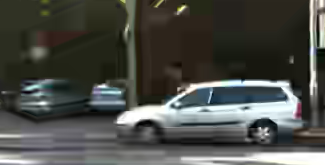} &
\includegraphics[width=0.2\linewidth]{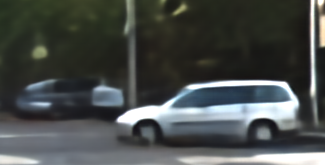} &
\includegraphics[width=0.2\linewidth]{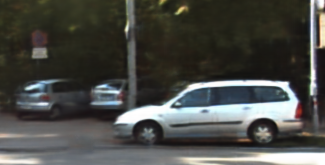} \\
0.04567 bpp & 0.04971 bpp & 0.04564 bpp & 0.04560 bpp \\


\includegraphics[width=0.2\linewidth]{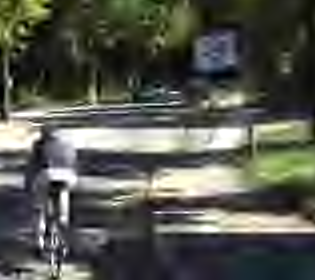} &
\includegraphics[width=0.2\linewidth]{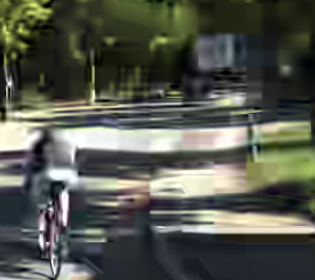} &
\includegraphics[width=0.2\linewidth]{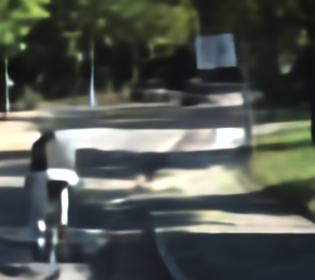} &
\includegraphics[width=0.2\linewidth]{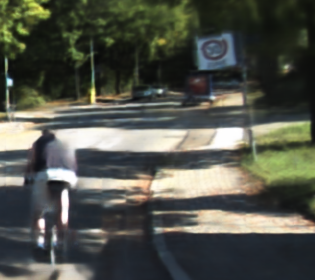} \\
0.06215 bpp & 0.06244 bpp & 0.05613 bpp & 0.05711 bpp \\


\end{tabular}
\caption{Our suggested method compared with the baseline model, JPEG 2000, and BPG over {\em KITTI General}.}

\label{fig:supp_examples_general_all4}
\end{center}
\end{figure*}






\begin{figure*}[htb!]
\begin{center}
\begin{tabular}{c}

\textbf{Without SI} 0.03437 bpp \\
\includegraphics[width=\linewidth]{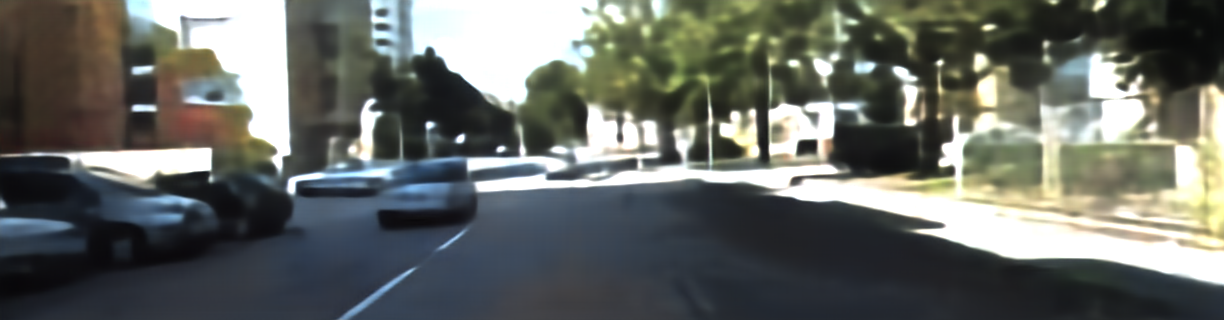} \\

\textbf{With SI} 0.03082 bpp \\
\includegraphics[width=\linewidth]{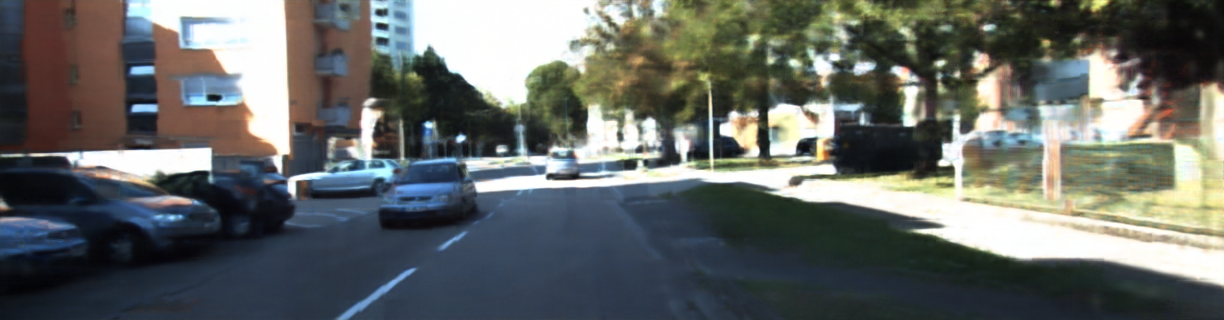} \\

\textbf{Without SI} 0.02608 bpp \\
\includegraphics[width=\linewidth]{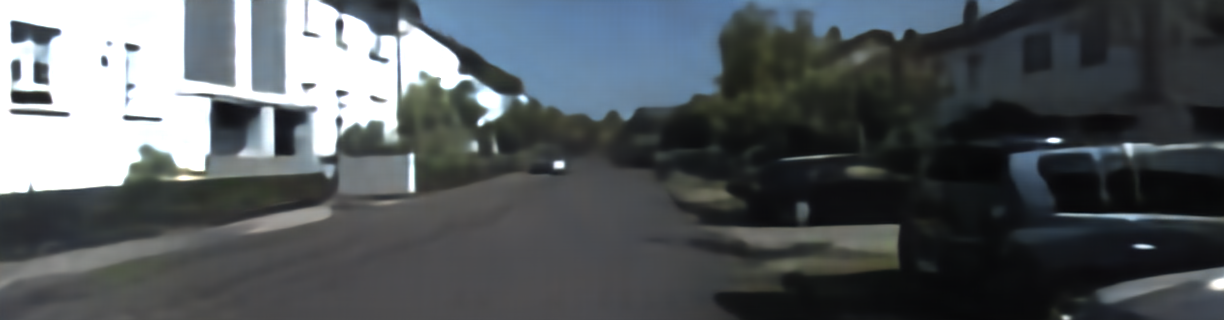} \\

\textbf{With SI} 0.02487 bpp \\
\includegraphics[width=\linewidth]{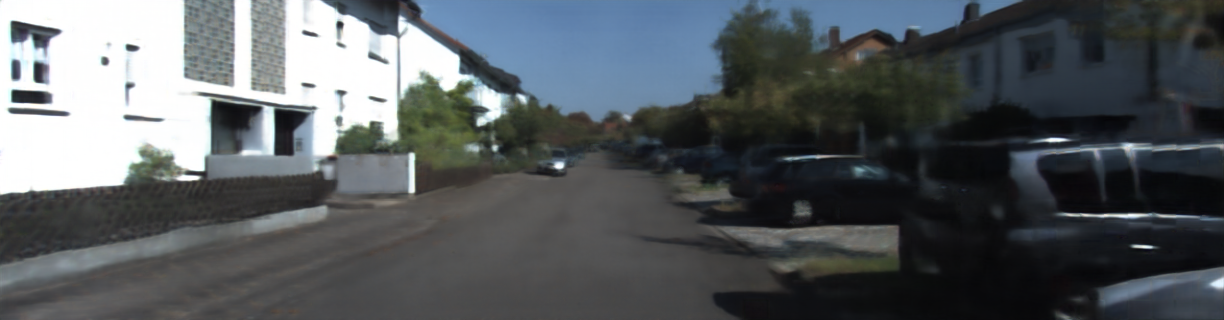} \\

\end{tabular}
\caption{Reconstruction comparison in low bit rates (that BPG failed to reach) over {\em KITTI Steereo} with and without side information.}

\label{fig:supp_examples_stereo_fullsize}
\end{center}
\end{figure*}

\begin{figure*}[htb!]
\begin{center}
\begin{tabular}{c}

\textbf{Without SI} 0.03931 bpp \\
\includegraphics[width=\linewidth]{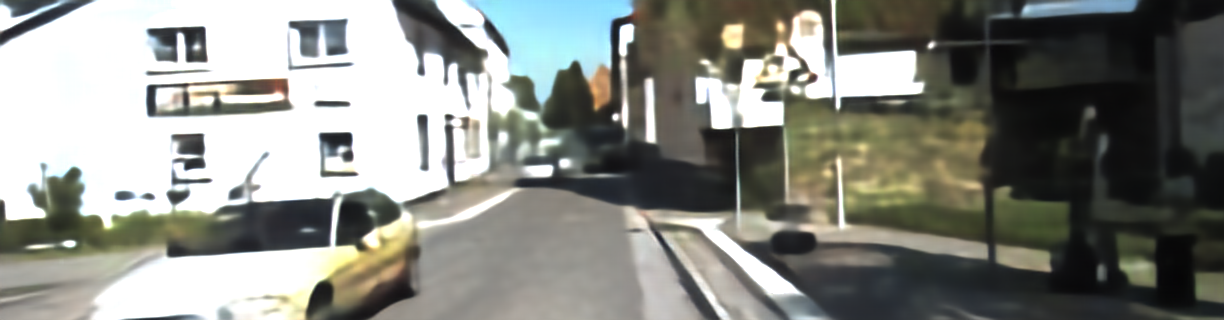} \\

\textbf{With SI} 0.03693 bpp \\
\includegraphics[width=\linewidth]{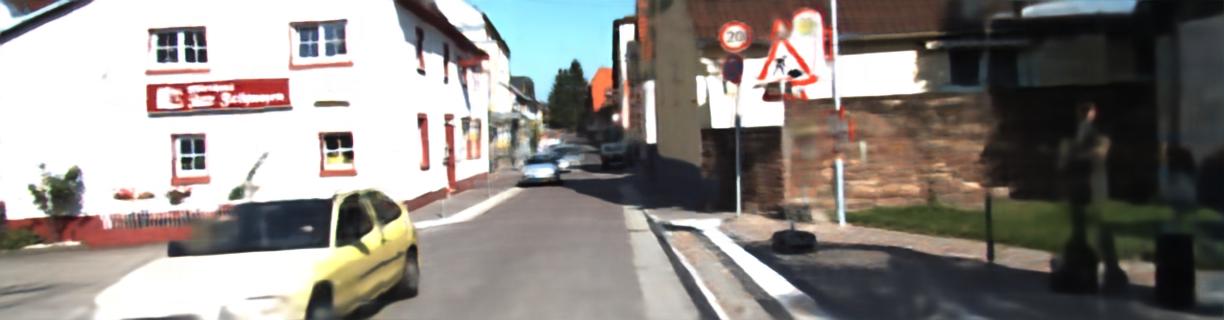} \\

\textbf{Without SI} 0.03105 bpp \\
\includegraphics[width=\linewidth]{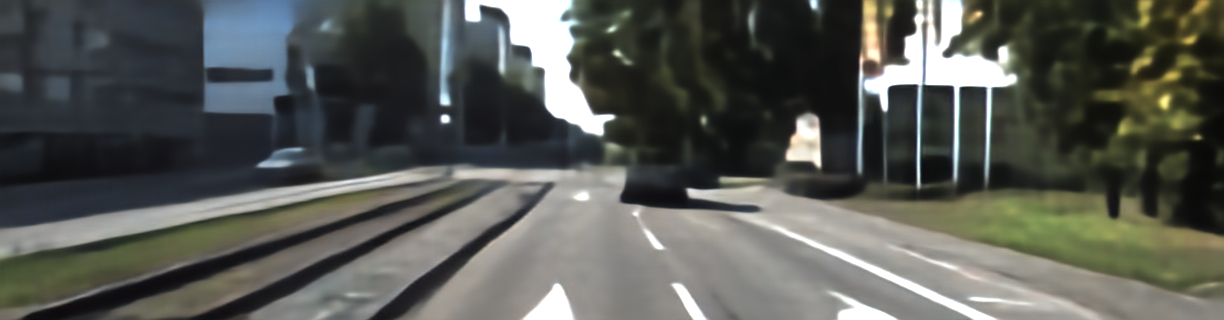} \\

\textbf{With SI} 0.02874 bpp \\
\includegraphics[width=\linewidth]{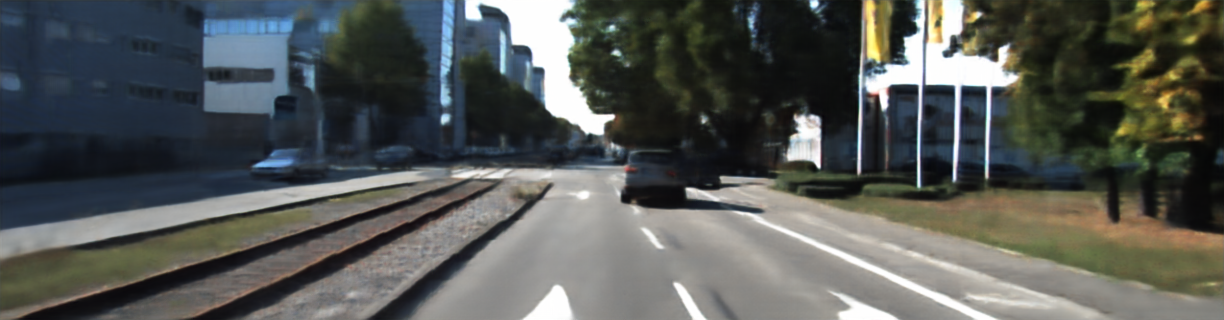} \\

\end{tabular}
\caption{Reconstruction comparison in low bit rates (that BPG failed to reach) over {\em KITTI General} with and without side information.}

\label{fig:supp_examples_general_fullsize}
\end{center}
\end{figure*}

\begin{figure*}[htb!]
\begin{center}
\begin{tabular}{c c}

\textbf{Without SI} 0.02789 bpp & \textbf{With SI} 0.02644 bpp\\
\includegraphics[width=0.4\linewidth]{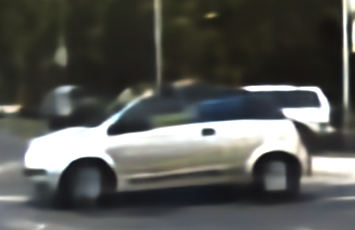} &
\includegraphics[width=0.4\linewidth]{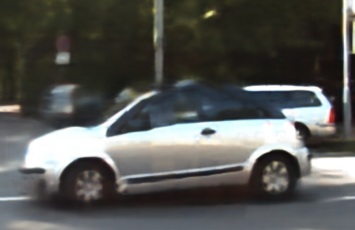} \\

\textbf{Without SI} 0.03377 bpp & \textbf{With SI} 0.03164 bpp \\ 
\includegraphics[width=0.4\linewidth]{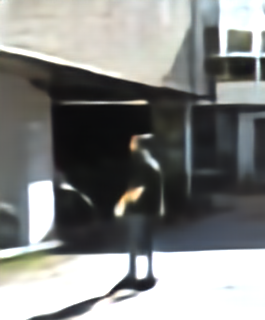} &
\includegraphics[width=0.4\linewidth]{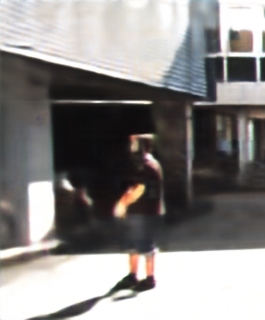} \\

\textbf{Without SI} 0.02601 bpp & \textbf{With SI} 0.02536 bpp \\
\includegraphics[width=0.4\linewidth]{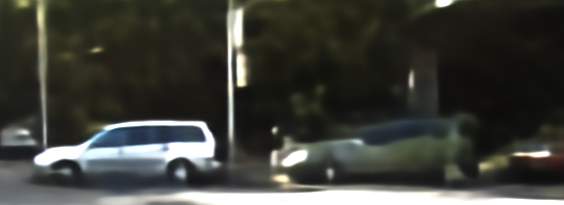} &
\includegraphics[width=0.4\linewidth]{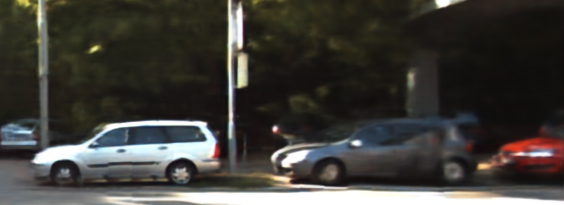} \\

\end{tabular}
\caption{Reconstruction comparison in low bit rates (that BPG failed to reach) over {\em KITTI General} with and without side information.}

\label{fig:supp_examples_general}
\end{center}
\end{figure*}

\begin{figure*}[htb!]
\begin{center}
\begin{tabular}{c c}

\textbf{Without SI} 0.03133 bpp & \textbf{With SI} 0.02981 bpp \\
\includegraphics[width=0.4\linewidth]{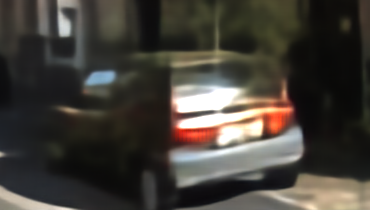} &
\includegraphics[width=0.4\linewidth]{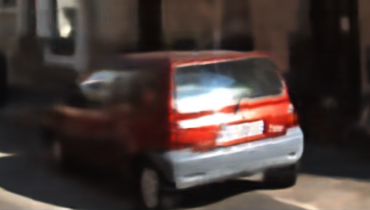} \\

 \textbf{Without SI} 0.02786 bpp & \textbf{With SI} 0.02760 bpp \\
\includegraphics[width=0.4\linewidth]{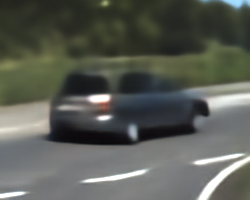} &
\includegraphics[width=0.4\linewidth]{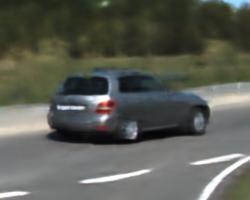} \\

\textbf{Without SI} 0.03461 bpp & \textbf{With SI} 0.03228 bpp \\
\includegraphics[width=0.4\linewidth]{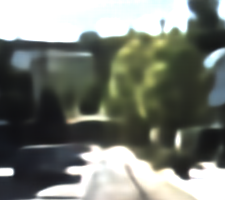} &
\includegraphics[width=0.4\linewidth]{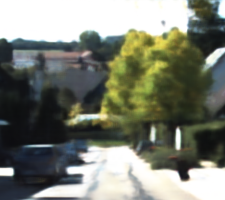} \\




\end{tabular}
\caption{Additional reconstruction comparison in low bit rates (that BPG failed to reach) over {\em KITTI General} with and without side information.}

\label{fig:supp_examples_general_cont}
\end{center}
\end{figure*}

\end{document}